\pdfoutput=1
\documentclass[11pt]{article}

\usepackage{EMNLP2023}

\usepackage{times}
\usepackage{latexsym}

\usepackage[T1]{fontenc}
\usepackage[utf8]{inputenc}
\usepackage{multirow}
\usepackage{microtype}

\usepackage{inconsolata}

\usepackage{graphicx}
\usepackage{booktabs}
\usepackage{subfig}
\usepackage{amsmath}
\usepackage{mdframed}
\definecolor{teaserblue}{RGB}{242, 242, 255}
\newcommand{\HOne}{$\mathcal{H}_1$}

\newcommand{\HTwo}{$\mathcal{H}_2$}
\usepackage{ifthen}
\newboolean{showcomments}
\setboolean{showcomments}{false} % Set to true to display comments, false to hide them

\newcommand{\SideNote}[2]{\ifthenelse{\boolean{showcomments}}{\todo[color=#1,size=\small]{#2}}{}} 
\newcommand{\deli}[1]{\ifthenelse{\boolean{showcomments}}{\textcolor{blue}{#1 - deli}}{}} 
\newcommand{\tux}[1]{\ifthenelse{\boolean{showcomments}}{\SideNote{purple!40}{#1}}{}} 
\newcommand{\lei}[1]{\ifthenelse{\boolean{showcomments}}{\textcolor{orange}{[#1 - lei]}}{}} 
\newcommand{\wang}[1]{\ifthenelse{\boolean{showcomments}}{\textcolor{red}{[#1 - wang]}}{}}

\title{Label Words are Anchors: An Information Flow Perspective for Understanding In-Context Learning}

\author{
Lean Wang$^{\dagger,\S}$,
Lei Li$^\dagger$, Damai Dai$^\dagger$, Deli Chen$^\S$, \\
\textbf{Hao Zhou$^\S$, Fandong Meng$^\S$, Jie Zhou$^\S$, Xu Sun$^\dagger$}\\ 
    $^\dagger$National Key Laboratory for Multimedia Information Processing,\\
      School of Computer Science, Peking University \\ 
      $^\S$Pattern Recognition Center, WeChat AI, Tencent Inc., China
      \\
 \texttt{\{lean,daidamai,xusun\}@pku.edu.cn} \quad
 \texttt{nlp.lilei@gmail.com} \\
 \texttt{victorchen@deepseek.com} \quad
 \texttt{\{tuxzhou,fandongmeng,withtomzhou\}@tencent.com}
  }

\begin{document}
\maketitle
\begin{abstract}
In-context learning (ICL) emerges as a promising capability of large language models (LLMs) by providing them with demonstration examples to perform diverse tasks. 
However, the underlying mechanism of how LLMs learn from the provided context remains under-explored. 
% What is information flow?
In this paper, we investigate the working mechanism of ICL through an information flow lens. 
Our findings reveal that label words in the demonstration examples function as anchors: (1) semantic information aggregates into label word representations during the shallow computation layers' processing; (2) the consolidated information in label words serves as a reference for LLMs' final predictions. Based on these insights, we introduce an anchor re-weighting method to improve ICL performance, a demonstration compression technique to expedite inference, and an analysis framework for diagnosing ICL errors in GPT2-XL. The promising applications of our findings again validate the uncovered ICL working mechanism and pave the way for future studies.\footnote{\url{https://github.com/lancopku/label-words-are-anchors}}

% \footnote{Our code is available at \url{https://github.com//Label}

\end{abstract}

\section{Introduction}
% Background 
In-context Learning (ICL) has emerged as a powerful capability alongside the development of scaled-up large language models (LLMs)~\citep{Brown2020LanguageMA}. 
By instructing LLMs using few-shot demonstration examples, ICL enables them to perform a wide range of tasks, such as text classification~\citep{Min2021NoisyCL} and mathematical reasoning~\citep{Wei2022ChainOT}. Since ICL does not require updates to millions or trillions of model parameters and relies on human-understandable natural language instructions~\citep{icl_survey}, it has become a promising approach for harnessing the full potentiality of LLMs. 
Despite its significance, the inner working mechanism of ICL remains an open question, garnering considerable interest from research communities~\citep{Xie2021AnEO, dai2022can, Akyrek2022WhatLA, Li2023TransformersAA}.

% Despite its significance, the inner working mechanism of ICL remains an open question, garnering considerable interest from research communities~\citep{Xie2021AnEO,Oswald2022TransformersLI,dai2022can,olsson2022inductionhead,Akyrek2022WhatLA,Garg2022WhatCT,Wang2023LargeLM,Tang2023LargeLM,Li2023TransformersAA}.

% \lei{add some recent studies as well ( later than April cannot excluded) -- have added some} % 加了

% 改成shallow deep layers,for GPT model（隐藏GPT2-xl这个具体的细节）

\begin{figure}[t!]
\centering
\includegraphics[width=0.45\textwidth]{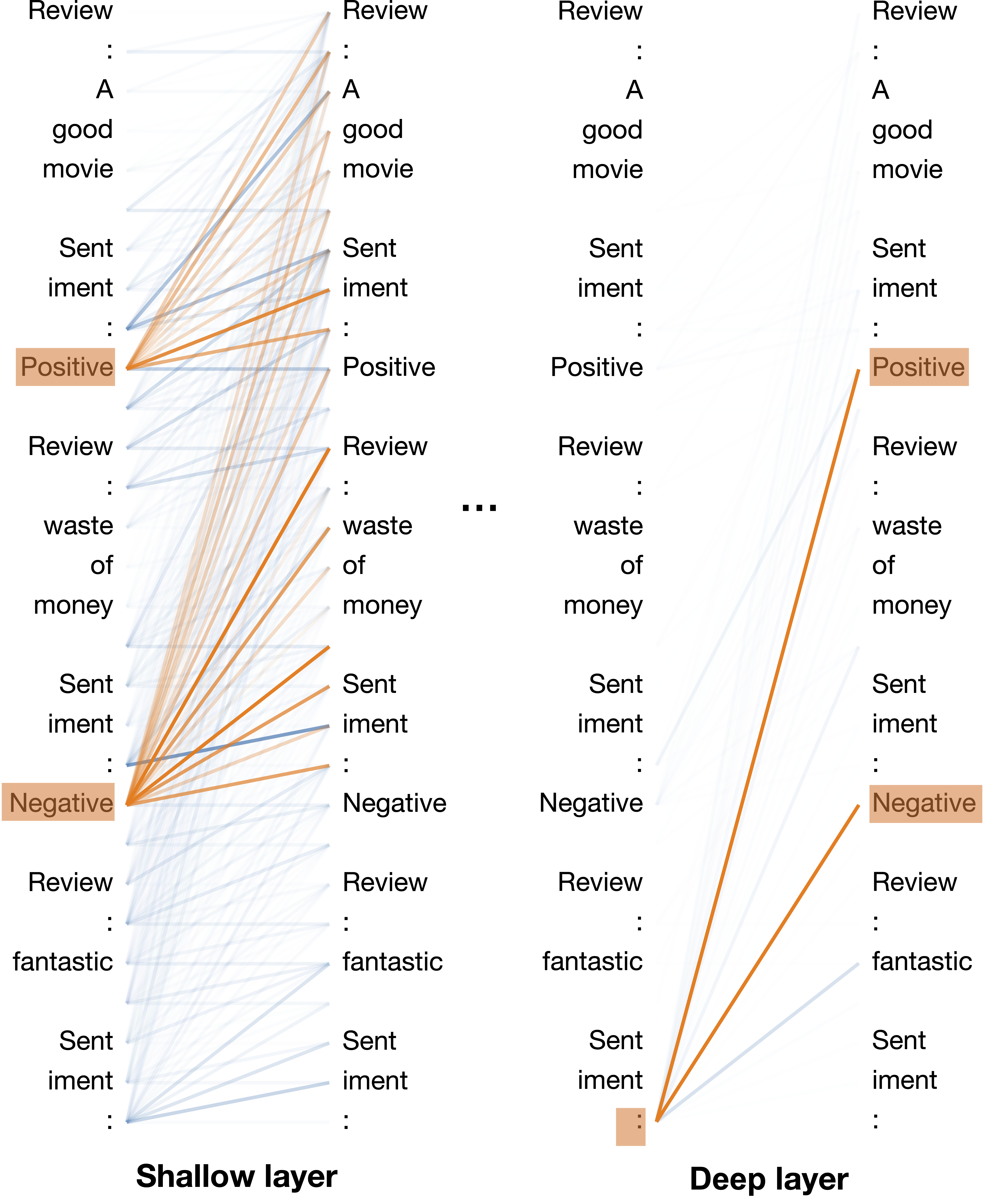}
\caption{
% \lei{redraw this figure to highlight the role of label words? - I plot a new version}
Visualization of the information flow in a GPT model performing ICL. The line depth reflects the significance of the information flow from the right word to the left. The flows involving label words are highlighted. Label words gather information from demonstrations in shallow layers, which is then extracted in deep layers for final prediction. }

\label{fig:vis}
\end{figure}

% % Motivation
% In this paper, we find that the label words serve as an anchor to aggregate and distribute information in ICL.
% Specifically, 
% % we explore the attention interactive pattern between tokens to gain an intuitive understanding of the information flow in ICL.
% % inspired by the fact that the attention mechanism is a core component  in the Transformer~\citep{Vaswani2017AttentionIA},
% we visualize the attention interactive pattern between tokens of predictions in GPT2-XL when performing a sentiment analysis task.
% As shown in Figure~\ref{fig:vis}, we intuitively notice that label words in demonstrations receive different attention patterns in shallow and deep Transformer layers, respectively, indicating that the information flow of ICL potentially follows a two-stage schema with label words as an anchor. 
% This intuition is further concretized by quantitative metrics based on saliency score, motivating us to propose a hypothesis on the ICL working mechanism:

% Motivation
% \deli{Why GPT2-XL? Need an explanation for this selection}
\begin{figure*}[t!]
\centering
\includegraphics[width=\textwidth]{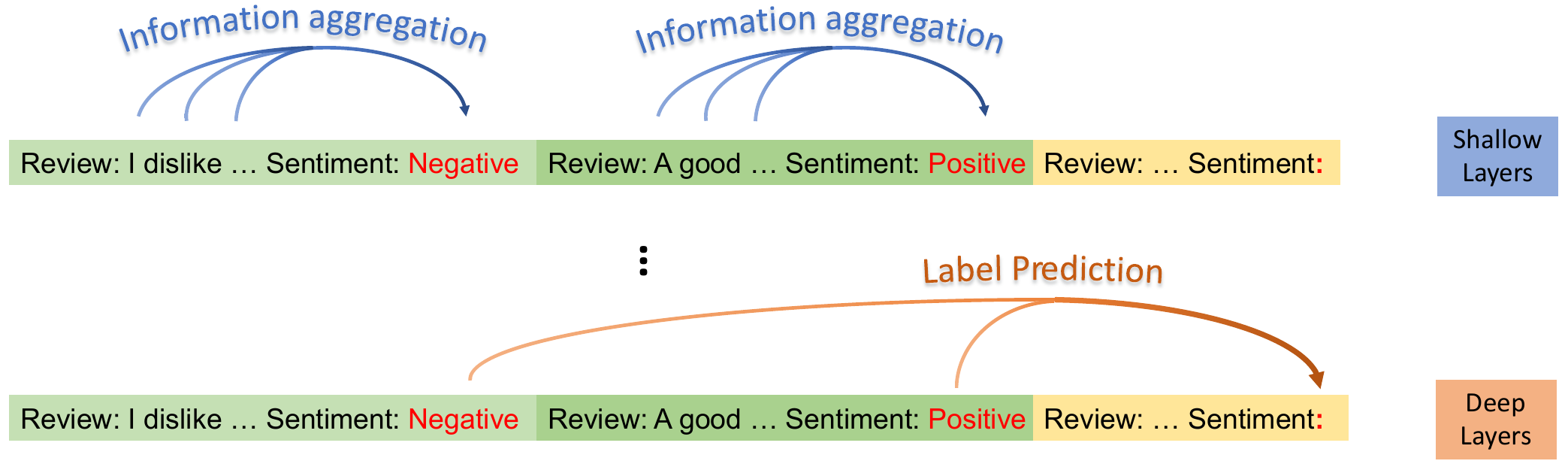}
\caption{Illustration of our hypothesis. In shallow layers, label words gather information from demonstrations to form semantic representations for deeper processing, while deep layers extract and utilize this information from label words to formulate the final prediction.
% \lei{explain what happens in the shallow/deep layers here}
}
\label{fig:icl_diagram}
\end{figure*}
In this paper, we find that the label words serve as anchors that aggregate and distribute information in ICL.
We first visualize the attention interactive pattern between tokens with a GPT model~\citep{Brown2020LanguageMA} on sentiment analysis (Figure~\ref{fig:vis}). Initial observations suggest that 
 label words aggregate information in shallow layers and distribute it in deep layers.\footnote{In this paper, ``shallow'' or ``first'' layers refer to those closer to the input, while ``deep'' or ``last'' layers are closer to the output. Here, ``deep layers'' include those around the midpoint, e.g., layers 25-48 in a 48-layer GPT2-XL.}  % ~\deli{Add a footnote to tell which layers are deep/shallow ?}
To draw a clearer picture of this phenomenon, 
we design two metrics based on saliency scores to portray the information flow in ICL and further propose the following hypothesis: 
\begin{mdframed}[backgroundcolor=teaserblue,hidealllines=true]
\noindent\emph{Information Flow with Labels as Anchors}

\noindent \HOne: In shallow layers, label words gather the information of demonstrations to form semantic representations for deeper layers. 

\noindent \HTwo: In deep layers, the model extracts the information from label words to form the final prediction.
\end{mdframed}

% To justify the hypothesis, 
% we first propose two metrics based on saliency scores to measure the influencing impact of label words in different parts of the models.

% quantitatively
% quantitatively
Two experiments are designed to validate the hypothesis using GPT2-XL~\citep{Radford2019LanguageMA} and GPT-J~\citep{gpt-j} across several text classification benchmarks.
% isolating 能不能换成更具体一点的词 这有点抽象 怎么 isolate ? disable attention? 
% blind? 
% By preventing the label words from attending to other words, so as to block ...
(1) By blocking the information aggregation path to label words in certain layers, we find that such isolation in shallow layers significantly impairs model performance. 
This indicates that label words collect useful information during forward propagation in shallow layers. 
(2) We investigate the relationship between the attention distributions on the label words of the target position and the model's final prediction. Our results illustrate a strong positive correlation, where a candidate label's probability increases with more attention weight on its corresponding label token.
% Our results show that the prediction positively correlates with the attention weights on label words, i.e., the probability of a candidate label would be higher with more attention weights on the specific label anchor. % 这么说可以吗? 只说 correlates 有点不好理解 需要解释一句 但不确定现在的解释是否符合实验？ 
% 应该符合的
% OK
% we study the correlation between the target position's attention of label words and model's prediction, and find that the correlation is strong on deeper layers. 
In summary, these experimental findings suggest that our hypothesis holds well with large language models on real-world datasets.
% So, the model decision process of ICL can be viewed as a two-stage process based on label words: (1) gather the information of demonstrations to label words (2) extract the information from label words to generate the final prediction.

% Drawing on insights from the information flow perspective, we investigate three approaches to enhance ICL's effectiveness, efficiency, and interpretability.
% First, we introduce an anchor re-weighting method utilizing a learnable vector to adaptively adjust the significance of various label words in demonstration examples, achieving a 7.8 average accuracy improvement over robust ICL baselines.
% Second, to expedite ICL inference, we compress its input into pre-computed anchor representations, as model predictions primarily rely on label word activations. Experiments demonstrate a threefold increase in speed with a minimal performance impact.
% Lastly, we present an error analysis example using ICL in GPT2-XL, revealing that the label confusion matrix closely mirrors the distance distribution of anchor key vectors, suggesting errors may arise from indistinguishable anchor representations. 
% These promising applications further validate our hypothesis and inform future ICL research.

% With the insights gained from the information flow perspective,
Drawing on insights from the information flow perspective, we explore three approaches to enhance ICL's effectiveness, efficiency, and interpretability.
(1) An anchor re-weighting method is introduced, which employs a learnable vector to adjust the significance of different label words in demonstrations, leading to a 16.7\% average accuracy boost compared to standard ICL baselines.
% average 1-shot v.s. re-weight 1-shot
% First, we propose an anchor re-weighting method that introduces a learnable vector to adaptively adjust the contribution of different label words in demonstration examples.
% Our method achieves xxx improvements over strong ICL baselines.
% Second, since the model predictions mainly depend on the activations of label words, we compress the original ICL input to pre-computed anchor representations to accelerate the inference of ICL.
(2)  For quicker ICL inference, inputs are compressed into pre-calculated anchor representations since model predictions primarily rely on label word activations. Testing shows a 1.8 $\times$ speedup in inference with only a minimal performance trade-off.
% 1.8 is the average over all datasets and two models
(3) An error analysis of ICL on GPT2-XL demonstrates that the label confusion matrix aligns closely with the distance distribution of anchor key vectors, implying that errors might result from similar anchor representations.
% Lastly, we showcase an error analysis example of ICL in GPT2-XL.
% We find that the label confusion matrix is highly similar to the distance distribution of the key vectors of anchors, indicating that the error potentially comes from the indistinguishable anchor representations.
These promising applications further validate our hypothesis and shed light on future ICL studies for better transparency of LLMs.
% Experiments show that our method can improve ICL's performance and efficiency, which further verifies our discovery, and points out new directions to improve ICL.
% saliency 图先放 introduction

% 只画 15-35,中间身略号,在删掉部分句子,短一点

% background icl火,作用机理依旧是开放性的问题,很多人在研究,我们也希望探索这个问题
% % 现在的分析工作不足；（这个不要了）
% 为了研究信息流动的作用,观察saliency map;
% 通过观察salinecy map图,启发我们以标签词为基础做分析,提出了一系列指标做分析
% 根据这个做分析实验；
% 以及基于规律提出了应用

% intro（每一段第一句话总结下；思路顺起来,前后承接）

% related work放倒数第二章

% 可以用New bing

% 要挂的话 section 5.23 前

% Organize the structure of the main paper 

\section{Label Words are Anchors}
\label{sec:anchor_hypothesis}
% 总起 说明这章做了什么 
This section confirms the intuitive findings using two saliency score-based metrics as discussed in \S~\ref{subsec:saliency_score}. The quantitative results lead to a proposed hypothesis for the ICL working mechanism:
\HOne: In shallow layers, label words aggregate information from demonstration examples to form semantic representations for later computations.
\HTwo: In deep layers, the model makes predictions by extracting information from label words.
The validation for these hypotheses is presented in \S~\ref{subsec:shallow_layer} and \S~\ref{subsec:deep_layer}, respectively.

\subsection{Hypothesis Motivated by Saliency Scores}
\label{subsec:saliency_score}
% \paragraph{Metrics}
% \paragraph{Experiments}
% 我改了下这两个部分标题,这样和intro的逻辑更一致一点
% 先写一下和之前的information flow的相关性什么的
% 先写一下干什么,为什么要用saliency score,开头写一下它是个常用的解释性方法,不要标题直接是Saliency Scores

 % Setup 
 This section aims to discover the inherent patterns in the attention interaction between tokens for a GPT model. The saliency technique~\cite{Simonyan2013DeepIC}, a common interpretation tool, is employed for highlighting critical token interactions. Following common practice, we use the Taylor expansion \cite{Michel2019AreSH} to calculate the saliency score for each element of the attention matrix:
% 改成 Following common practice, we use
% Taylor expansion...
\begin{equation}
\small 
I_{l}=\left|\sum_h A_{h,l} \odot \frac{\partial \mathcal{L}(x)}{\partial A_{h,l}}\right|.
\end{equation}
Here, $A_{h,l}$ is the value of the attention matrix of the $h$-th attention head in the $l$-th layer, $x$ is the input, and $\mathcal{L}(x)$ is the loss function of the task, e.g., the cross-entropy objective for a classification problem. We average all attention heads to obtain the saliency matrix $I_{l}$ for the $l$-th layer.\footnote{Another choice is to use $I_{l}=\sum_h\left|A_{h,l} \odot \frac{\partial \mathcal{L}(x)}{\partial A_{h,l}}\right|$, which raises quite similar results.}  %\tux{(i, j) seems contrary to Eq.2.3.4}
$I_{l}(i,j)$ represents the significance of the information flow from the $j$-th word to the $i$-th word for ICL.
By observing $I_{l}$, we can get an intuitive impression that as the layer goes deeper, demonstration label words will become more dominant for the prediction, as depicted in Figure~\ref{fig:vis}.
% Using $I_{l}$, we can get visualization results like Figure~\ref{fig:vis}, and the hypothesis that label words play an important role in ICL.

% 就写一下根据I猜想。。。,like figure 1(顺带提一下就行)

To draw a clearer picture of this phenomenon, we propose three quantitative metrics based on $I_{l}$. Our focus lies in three components: (i) the label words, such as ``Negative'' and ``Positive'' in Figure~\ref{fig:icl_diagram}, denoted as $p_1,...,p_C$, where $C$ represents the total number of label words;\footnote{In this study, the term 'label words' is approximately equal to 'label tokens'. The only deviation is the 'Abbreviation' in the TREC dataset, where we use the first subword in experiments, following \citet{Zhao2021CalibrateBU}.} (ii) the target position, where the model generates prediction labels (i.e., the final token in the input), which we denote as $q$; and (iii) the text part, i.e., the tokens before label words in the demonstration.

% \lei{target position should be specified here}
% \wang{Add explanations. Is it clear enough now?}
The definitions of the three quantitative metrics follow below.
% \lei{Chinglish here, the denotation of symbols should be paraphrased with ChatGPT  -- have paraphrased to a new version} % 改了下
%, which measures the average importance of the information flow of different parts:
% label word特殊,我们想看三种关系的信息流动；我们看三种关系的信息流动的强弱,所以有三个metrics
% where 不要,用align,两行

% \noindent\textbf{$\boldsymbol S_{wp}$, the average importance of the information flow from other words to label words:}
\noindent\textbf{$\boldsymbol S_{wp}$, the mean significance of information flow from the text part to label words:}
\begin{equation}
\small 
    \begin{aligned}
    S_{wp} &= \frac{\sum_{(i,j)\in C_{wp}}{I_l(i,j)}}{|C_{wp}|},  \\ 
    C_{wp} &= \{(p_k,j):k\in [1,C], j< p_k\}.
    \end{aligned}
\end{equation}
\noindent\textbf{$\boldsymbol S_{pq}$, the mean significance of information flow from label words to the target position:}%\tux{TODO}
\begin{equation}
\small 
    \begin{aligned}
    S_{pq} &= \frac{\sum_{(i,j)\in C_{pq}}{I_l(i,j)}}{|C_{pq}|}, \\ 
    C_{pq} &= \{(q,p_k):k\in [1,C]\}.
\end{aligned}
\end{equation}

\noindent\textbf{$\boldsymbol S_{ww}$, the mean significance of the information flow amongst all words, excluding influences represented by $\boldsymbol S_{wp}$ and $\boldsymbol S_{pq}$ :}%\tux{TODO}
% xxx for xxx:
\begin{equation}
\small 
    \begin{aligned}
         S_{ww} = &\frac{\sum_{(i,j)\in C_{ww}}{I_l(i,j)}}{|C_{ww}|},  \\ 
   C_{ww} =& \{(i,j): j<i\}- C_{wp} - C_{pq}.
    \end{aligned}
\end{equation}

% \lei{Eq 4 has two rows, but I cannot see the connection between the index and -Cwp - Cpq}
% \lei{add a brief sentence here to summarize the metric. xxx higher means xxx while xxx indicates xxx}

$S_{wp}$, $S_{pq}$, and $S_{ww}$ help assess different information flows in the model. $S_{wp}$ indicates the intensity of information aggregation onto label words. A high $S_{pq}$ demonstrates a strong information extraction from label words for final decision-making. $S_{ww}$ assesses average information flow among words, serving as a benchmark to gauge the intensity of the patterns identified by $S_{wp}$ and $S_{pq}$.
%不知道要不要解释一下j< p_k,忽略$I_l(i,i)$的问题,还是省掉
% Here, in the above three indicators, the $j$ in $I_l(i,j)$ is always less than $i$, because each word can only get information from words before it in generative language models like GPT2-XL. We also ignore $I_l(i, i)$, since it is the attention of word $i$ to itself, which does not reflect the interaction between tokens.

\paragraph{Experimental Settings}
\label{para:experimental_settings_hypothesis}
% 后面GPT-J加上的话再补充；说下为什么用GPT2-XL
We choose GPT2-XL from the GPT series~\cite{Radford2019LanguageMA} as our primary model for investigation, due to its moderate model size~(of 1.5B parameters) that is suitable for our hardware resource and its decent ICL performance~\citep{dai2022can}. 
% GPT2-XL is widely adopted in ICL works and enables us to calculate saliency scores on it. 
For datasets, we use Stanford Sentiment Treebank Binary~(SST-2)~\cite{socher-etal-2013-recursive} for sentiment analysis, Text REtrieval Conference Question Classification~(TREC)~\cite{li-roth-2002-learning,hovy-etal-2001-toward} for question type classification, AG's news topic classification dataset~(AGNews)~\cite{Zhang2015CharacterlevelCN} for topic classification, and EmoContext~(EmoC)~\cite{chatterjee-etal-2019-semeval} for emotion classification. Templates for constructing demonstrations are provided in Appendix~\ref{apx:dataset}.
$1000$ examples are sampled from the test set for evaluation, with one demonstration per class sampled from the training set. Experiments with more demonstrations yield similar outcomes (refer to Appendix~\ref{appendix:demonstration_2} for details). Results reflect averages from five random seeds.% We used GPT2-XL\footnote{Due to our computational resource constraints that prevent us from computing gradients on GPT-J, we omit GPT-J for analysis.} on each dataset using 5 random seeds, and for each random seed, we extracted 1000 inputs to be predicted from the test set\footnote{If the size of the test set (we use the validation set for SST-2) is less than 1000, we use the entire test set.}, and for each input to be predicted, we extracted one demonstration per category from the training set as the demonstration set (we extracted one demonstration per category for ease of analysis). 

% Model, why GPT2-XL
% Dataset,  statistics refer to Appendix 

\paragraph{Results and Analysis}
\label{para:saliency_score_ana}
% 核心结论加粗/斜体,标题可以删掉analysis,或者换hypothesis
\begin{figure}[t!]
\centering
\subfloat[Results on the SST-2 dataset]{
\label{sfig:attn_attr_sst2}
\includegraphics[height=5.5cm]{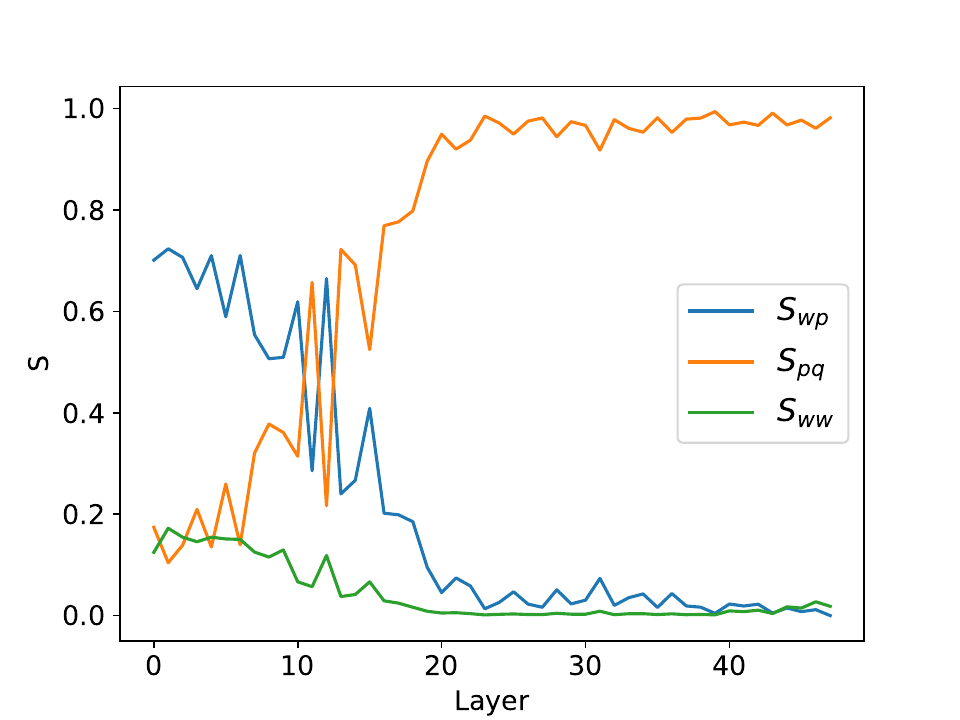}}
% \hspace{2em}
\hfill
\subfloat[Results on the AGNews dataset]{
\label{sfig:attn_attr_agnews}
\includegraphics[height=5.5cm]{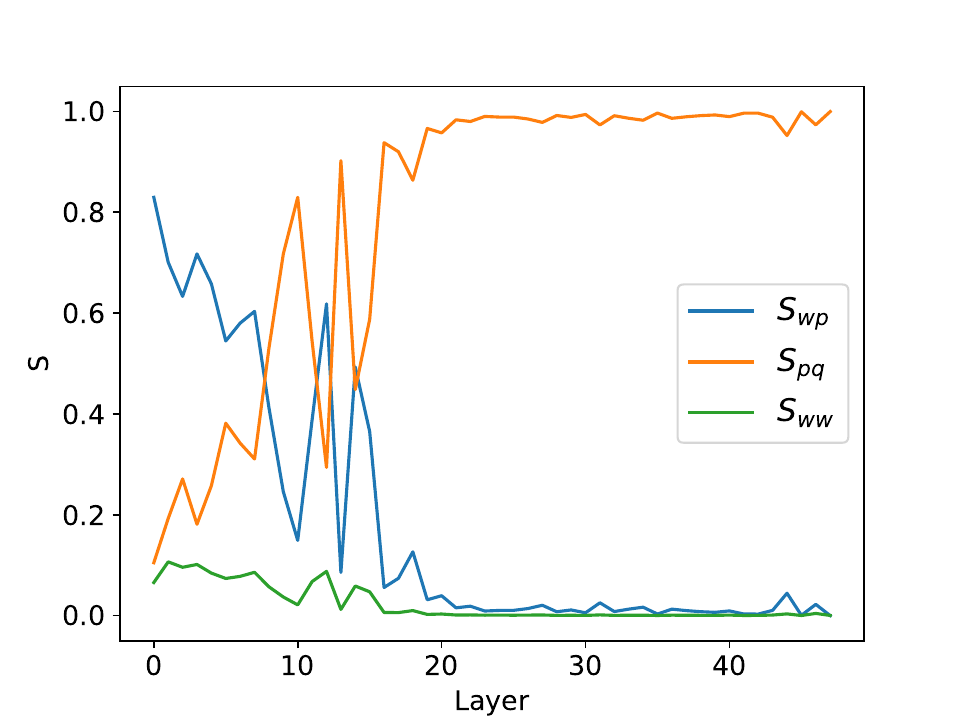}}

% \begin{figure}[t!]
% \centering
% \includegraphics[height=5.5cm]{photos/attn_attr_agnews_1.pdf}
\caption{Relative sizes of $S_{wp}$, $S_{pq}$, and $S_{ww}$ in different layers on SST-2 and AGNews. Results of other datasets can be found in Appendix~\ref{sec:appdendix_trec_emoc}. 
Initially, $S_{wp}$ occupies a significant proportion, but it gradually decays over layers, while $S_{pq}$ becomes the dominant one.}
\label{fig:attn_attr}

% attention_attr_ana.ipynb
\end{figure}

Figure~\ref{fig:attn_attr} reveals that: (1) in shallow layers, $S_{pq}$, the significance of the information flow from label words to targeted positions, is low, while $S_{wp}$, the information flow from the text part to label words is high; (2) in deep layers, $S_{pq}$, the importance of information flow from label words to the targeted position becomes the dominant one. Notably, $S_{pq}$ and $S_{wp}$ usually surpass $S_{ww}$, suggesting that interactions involving label words outweigh others.

\paragraph{Proposed Hypothesis}
% 这里要不要也弄个代码块
Based on this, we propose the hypothesis that label words function as anchors in the ICL information flow. In shallow layers, label words gather information from demonstration examples to form semantic representations for deeper layers, while in deep layers, the model extracts the information from label words to form the final prediction.
Figure~\ref{fig:icl_diagram} gives an illustration for our hypothesis.

\subsection{Shallow Layers: Information Aggregation}
% 我们从前面的可视化猜测信息流动规律,然后信息流动/token互联在transformer里是用attention实现,然后我们通过调整attention可以阻断信息流动。如果阻断信息流动后,模型效果变差了,就说明aggregation存在。
\label{subsec:shallow_layer}
% In this part, we validate the first part of our hypothesis. Specifically, we prevent the label words from attending to the demonstration contexts, to examine the existence and effect of information aggregation on label words.
In this part, we validate our hypothesis' first component. We assume that the information aggregation in ICL relies on the information flow from the
% \lei{what are text parts? maybe we should give a clear definition before and use it consistently.}
text part to label tokens, which is facilitated by the transformer's attention mechanism. 
By manipulating the attention layer in the model to block this flow and examining the model behavior change,
we validate the existence of the information aggregation process and its contribution to the final prediction.

% 加一个setting部分,LL/WL的metric放setting里 
\paragraph{Experimental Settings}
\label{para:experimental_settings_shallow}
We retain the same test sample size of 1000 inputs as \S~\ref{para:experimental_settings_hypothesis}. We use the same demonstration for a single random seed. To further validate our findings on larger models, we incorporate GPT-J~(6B)~\citep{gpt-j} in experiments, which exceeds GPT2-XL in model size and capacity. 

\paragraph{Implementation Details}
 % Setup 
 To block the information flow to label words, we isolate label words by manipulating the attention matrix $A$.
 Specifically, we set $A_{l}(p, i) (i<p)$  to 0 in the attention matrix $A_{l}$ of the $l$-th layer, where $p$ represents label words and $i$ represents preceding words. Consequently, in the $l$-th layer, label words cannot access information from the prior demonstration text.

\paragraph{Metrics}
% 把WL 为什么要提这个从footnote里放到正文
We use the following metrics to assess the impact of blocking information flow from the text part to label tokens:
% \footnote{The reason for adding word loyalty in addition to label loyalty is that it can show more information in some cases. See detailed discussion in Appendix~\ref{app:sec_wl_ll}}
\textbf{(1) Label Loyalty:} measures the consistency of output labels with and without isolation.  
\textbf{(2) Word Loyalty:} employs the Jaccard similarity to compare the top-5 predicted words with and without isolation, capturing more subtle model output alterations (See Appendix~\ref{app:sec_wl_ll} for details).
Low loyalty indicates a profound impact of isolation on model predictions.

\begin{figure}[t!]
    \centering
    \includegraphics[width=0.45\textwidth]{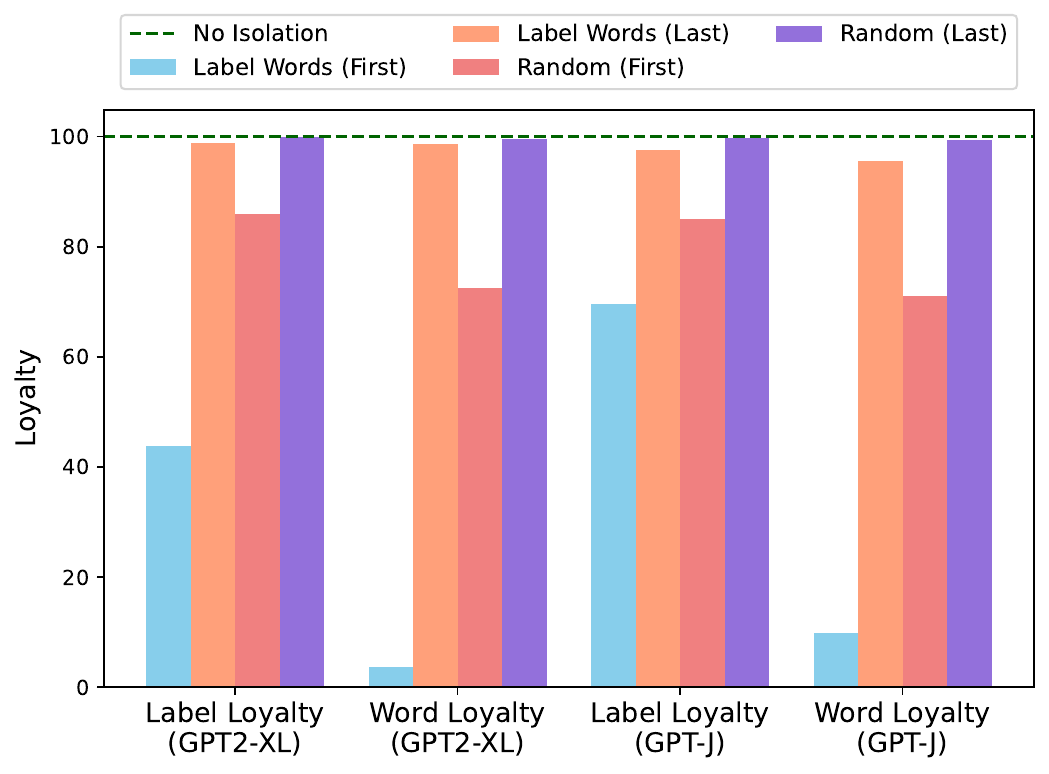}
    \caption{ The impact of isolating label words versus randomly isolating non-label words within the first or last 5 layers. Isolating label words within the first 5 layers exerts the most substantial impact, highlighting the importance of shallow-layer information aggregation via label words.}
    \label{fig:agg_result}
\end{figure}

\paragraph{Results and Analysis}
% 浅层阻断有影响说明aggregation存在,深层影响小说明aggregation在浅层

Figure~\ref{fig:agg_result} illustrates a notable influence on the model's behavior when label words are isolated within the first 5 layers. Yet, this influence becomes inconsequential within the last 5 layers, or when random non-label words are used. This observation underlines the fundamental importance of shallow-layer information aggregation via label words in ICL. It also emphasizes the superiority of label words over non-label words. Further tests with variable numbers of layers reaffirm these findings (Appendix~\ref{appendix:isolating_different_num}). Moreover, similar results were obtained when testing ICL with semantically unrelated labels (refer to Appendix~\ref{app:SULICL}).

\subsection{Deep Layers: Information Extraction}

% 好像把单层的图删了只画累计和的话,这样子AUCROC本身的值看不出来,没法argue相关性,只能说后几层更重要,还是得加回去？用透明图叠加一下,画四个task平均

% 说一下前面的发现,然后为了验证发现,我们衡量了correlation
\label{subsec:deep_layer}
We proceed to validate the latter part of our hypothesis that the model extracts information from label words to form the final
prediction. We denote the sum of the attention matrices in the $l$-th layer as $A_l$.\footnote{Here we sum up the attention matrices of all attention heads in the $l$th layer for convenience of analysis.} In deeper layers, we find a strong correlation between the attention distributions on the label words of the target position, represented as $(A_l(q,p_1),..., A_l(q,p_C))$, and the model's final prediction, affirming our hypothesis. 
% This is analyzed through the correlation between the attention distributions on label words of the target position and the model's final prediction.
% \lei{why analysis this way? add a sentence to explain the motivation here. i.e., adjust the first paragraph of 2.3.1 to here and adjust it accordingly}
The experimental setup mirrors that discussed in \S~\ref{para:experimental_settings_shallow}.

% 把 metrics提出来,AUC-ROC的计算简短一些
\subsubsection{Experiments}
 % Setup 
 % \lei{we have defined these before? remove it if it is clear to understand this notation}
% Suppose the sum of the attention matrices of the $l$-th layer is $A_l$,\footnote{Here we sum up the matrices of all attention heads in a layer for convenience of analysis} we postulate that for deep layers, there is a strong correlation between the attention distributions on the label words of the target position, denoted as $(A_l(q,p_1),..., A_l(q,p_C))$, and
% the model's final prediction. 

We utilize the AUC-ROC score to quantify the correlation between $A_l(q,p_i)$ and model prediction, which we denote as $\text{AUCROC}_l$ for the $l$-th layer. We prefer the AUC-ROC metric due to two primary reasons: (1) $A_l(q,p_i)$ might differ from the probability of the model outputting label $i$ by a constant factor. As \citet{Kobayashi2020AttentionIN} points out, attention should be multiplied by the norm of the key vector to yield 'more interpretable attention'. The AUC-ROC metric can implicitly account for these factors, thus allowing us to uncover the correlation more effectively. (2) The proportion of different labels output by the model may be unbalanced. Using the AUC-ROC metric can help mitigate this issue, reducing disturbances caused by class imbalance.
 % \lei{the first reason is not clear to me? maybe paraphrase it or elaborate more on this?}
 % \wang{Is it clearer now?}

% Besides, it is worth noting that the AUC-ROC values of each layer here reflect the contribution of each layer to the final classification result, rather than the cumulative contribution from the beginning to that layer. For example, in Figure \ref{fig:auc-roc-gpt2-xl}, the AUC-ROC values corresponding to the last few layers of SST-2 are low, but this does not mean that the classification information hidden in the positions to be classified in these layers does not match the final output. Instead, it means that the model has already completed the classification task by utilizing the attention to the label word positions at the prediction positions, and these layers no longer need to continue to pay attention to the label word positions.

% To measure the relationship between the classification information hidden in the positions to be classified in each layer and the final output, we need to further process the above metrics. 
Considering the residual mechanism of transformers, we can view each layer's hidden state as the cumulative effect of all prior layer calculations. To quantify the accumulated contribution of the first $l$ layers to model prediction, 
% Therefore, to reflect the cumulative classification information extracted from label words before the $l$th layer, we can define the following measure:
we introduce $R_l$:
\begin{equation}
\small 
R_l = \frac{\sum_{i=1}^l (\text{AUCROC}_i -0.5)}{\sum_{i=1}^N (\text{AUCROC}_i-0.5)}.
\end{equation}
This measure tracks the positive contribution above a baseline AUC-ROC threshold of 0.5. The value of $R_l$ signifies the proportional contribution of the first $l$ layers to the model prediction.

\subsubsection{Results and Analysis}
\begin{figure}[t!]
  \centering
  \subfloat[GPT2-XL (total 48 layers).]{
    \includegraphics[width=0.45\textwidth]{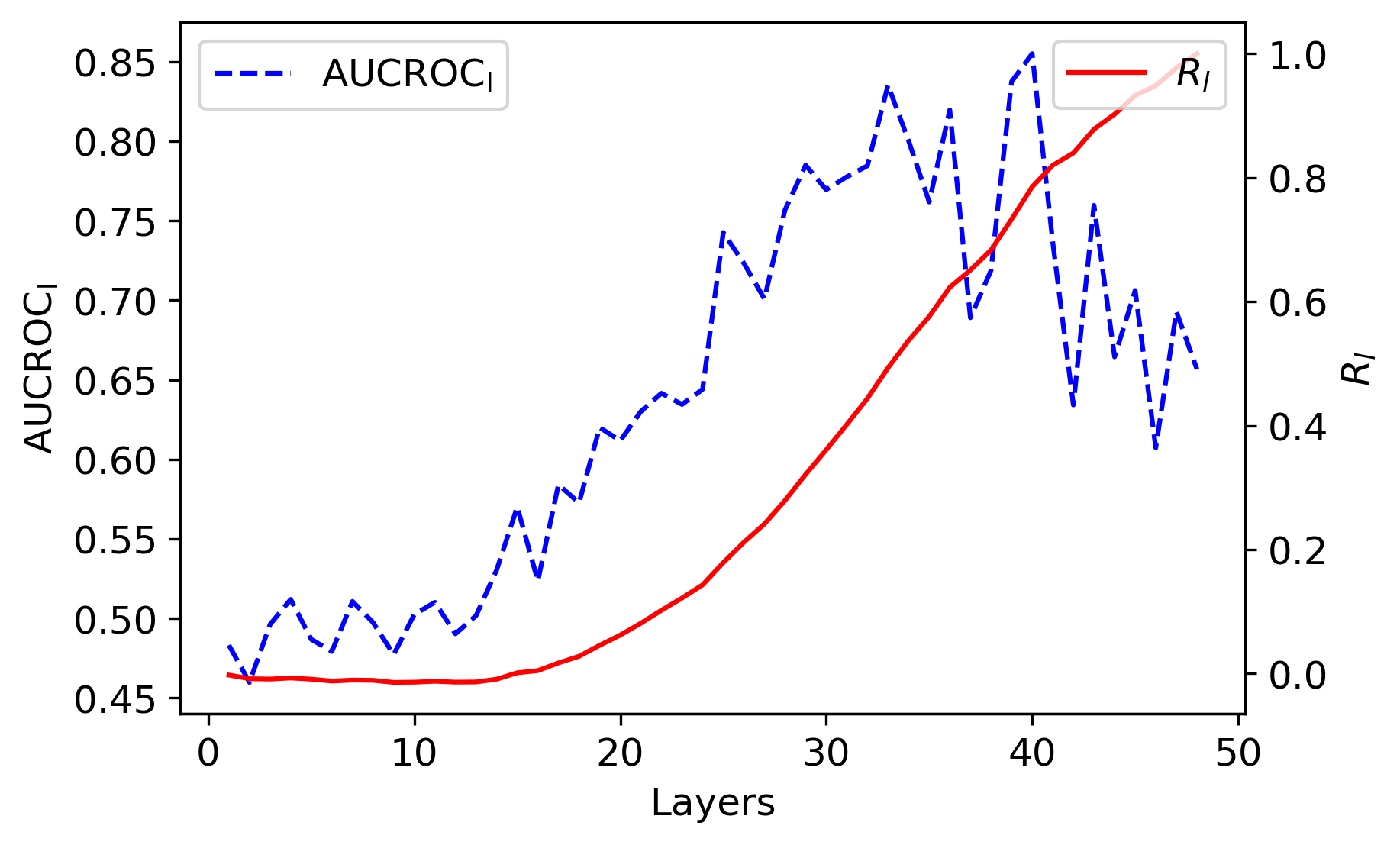}
    \label{fig:auc-roc-ratio-gpt2-xl}}
  \hfill
  \subfloat[GPT-J (total 28 layers).]{
    \includegraphics[width=0.45\textwidth]{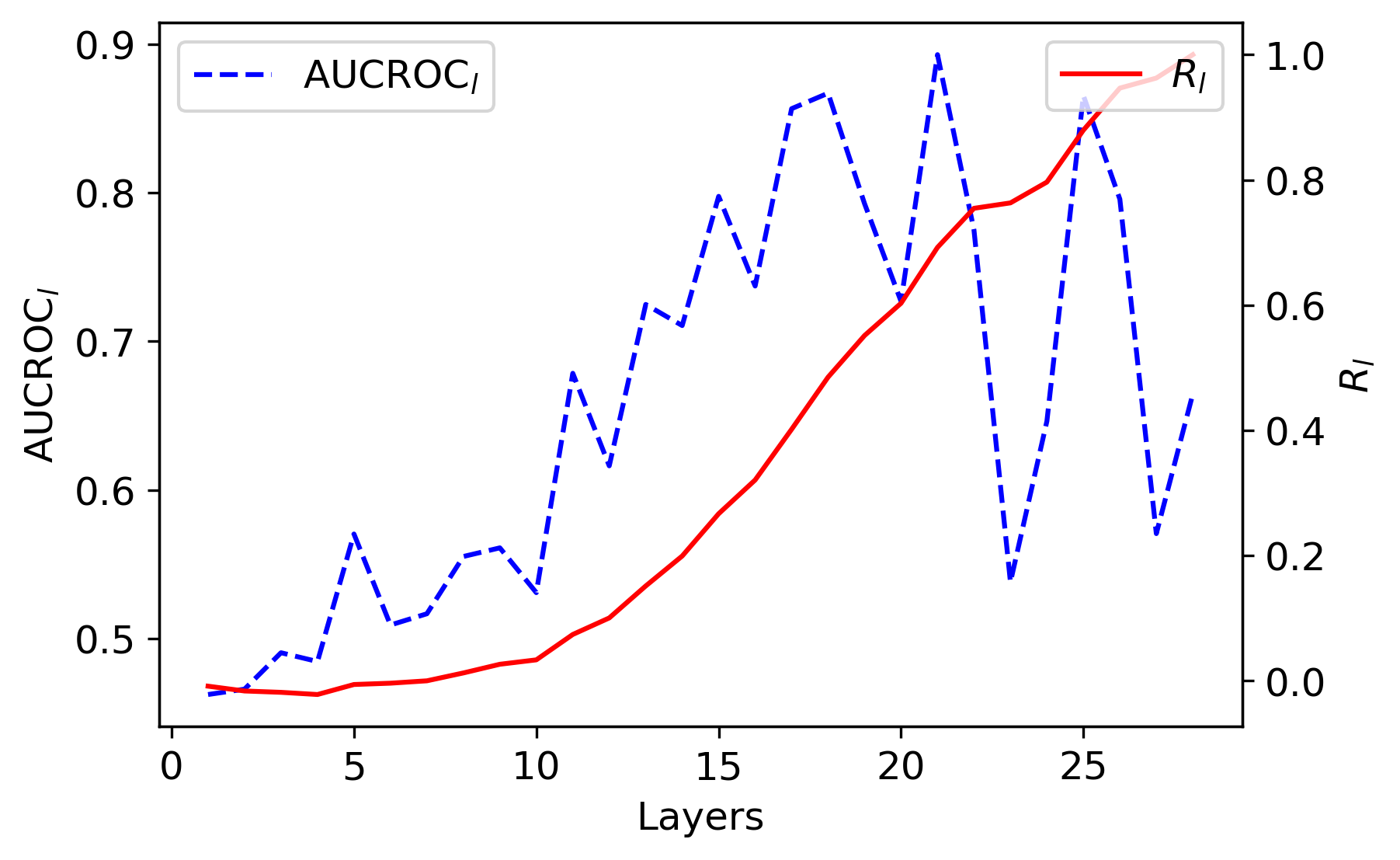}
    \label{fig:auc-roc-ratio-gpt-j}}
  \caption{$\text{AUCROC}_l$ and $R_l$ of each layer in GPT models. The result is averaged over SST-2, TREC, AGNews, and Emoc. $\text{AUCROC}_l$ reaches 0.8 in deep layers, and $R_l$  increases mainly in the middle and later layers.% add conclusion
  } 
  \label{fig:auc-roc-ratio}
\end{figure}

Figures \ref{fig:auc-roc-ratio-gpt2-xl} and \ref{fig:auc-roc-ratio-gpt-j} delineate correlation metrics for GPT2-XL and GPT-J, averaged across four datasets. 
The $\text{AUCROC}_l$ for deep layers approaches $0.8$, illustrating a strong correlation between the attention distributions on label
words of the target position and the model's final
prediction. Moreover, shallow layers show negligible cumulative contributions ($R_l$), with a significant increase in middle and deep layers.  These results signify the crucial role of deep layers for final prediction, validating that the model extracts information from label words in deep layers to form the final prediction.
% Figure \ref{fig:auc-roc-ratio-gpt2-xl} and Figure \ref{fig:auc-roc-ratio-gpt-j} show the correlation metrics of each layer for GPT2-XL and GPT-J, respectively. 
% % \lei{Figure 10? seem a wrong index}
% The result is averaged over four datasets. 
% Firstly, the $\text{AUCROC}_l$ of the deep layers reaches a high score of $0.8$, indicating a strong correlation between the attention distributions on the label
% words of the target position and the model's final
% prediction. Secondly, the cumulative contributions $R_l$ of the first few layers are near $0$, while the score increases significantly in the middle and later layers. This phenomenon demonstrates that the classification decision of the model mainly takes place in the deep layers. From these two points, we validate that the model extracts the information from label words in deep layers to form the final prediction.

% draft
\subsection{Discussion of Our Hypothesis}

In \S~\ref{subsec:shallow_layer}, we have affirmed that the model's shallow layers assemble information from demonstrations via label words to form semantic representations. In \S~\ref{subsec:deep_layer}, we verify that the aforementioned aggregated information on label words is then extracted to form the final prediction in the deep layers. Recognizing the crucial function of label words in this process, we have introduced the term ``Anchors'' to denote them. Given the considerable role these ``anchors'' fulfill, we find it intuitive to design ICL improvements based on them, as elaborated in \S~\ref{sec:applications}.

\section{Applications of Our Anchor-Based Understanding}
\label{sec:applications}

% interpretability of ICL.可以删了,因为不强
With insights from the validated hypothesis, we propose strategies to boost ICL's accuracy and inference speed. We propose an anchor re-weighting method in \S~\ref{subsec:anchor_re-weighting} to adjust the demonstrations' contributions and improve accuracy. 
In \S~\ref{subsec:compress}, we explore a context compression technique that reduces original demonstrations to anchor hidden states to speed up ICL inference. Besides, in \S~\ref{subsec:error_analysis}, we utilize anchor distances to perform an analysis to understand the errors ICL made in real-world scenarios.
These approaches corroborate our hypothesis, pointing to potential paths for future ICL enhancements.

\begin{table*}[t!]
\centering

\begin{tabular}{c|cccc|c}
\toprule
Method&  SST-2 & TREC & AGNews & EmoC & Average \\
\midrule
Vanilla In-Context Learning ( 1-shot per class ) & $61.28$ & $57.56$ & $73.32$ & $15.44$ & $51.90$ \\
Vanilla In-Context Learning ( 5-shot per class )& $ 64.75 $ & $ 60.40 $ & $ 52.52 $ & $ 9.80$ & $46.87$ \\
Anchor Re-weighting (1-shot per class) & $\textbf{90.07}$ & $\textbf{60.92}$ & $\textbf{81.94}$ & $\textbf{41.64}$ & $\textbf{68.64}$ \\
\bottomrule
\end{tabular}
\caption{The effect after adding parameter $\beta_0^i$. For AGNews, due to the length limit, we only use three demonstrations per class. Our Anchor Re-weighting method achieves the best performance overall tasks. }
\label{tab:lda_1}
\end{table*}

% \subsection{Weighted ICL: Linking Attention Mechanism to Logistic Regression}
\subsection{Anchor Re-weighting}
\label{subsec:anchor_re-weighting}
Based on our analysis in \S~\ref{sec:anchor_hypothesis}, we draw parallels between ICL and logistic regression and propose an approach to improve ICL's accuracy by re-weighting label anchors.

% 用subsubsection, paragraph只管一段

\subsubsection{Method}

\S~\ref{subsec:deep_layer} illustrates a strong correlation between the model's output category and the attention distribution $\left(A\left(q, p_1\right), \ldots, A\left(q, p_C\right)\right)$ on label words $p_1,...,p_C$ of the target position $q$ in deep layers. We can view the attention module as a classifier $\boldsymbol f$,
\begin{equation}
\small 
\begin{aligned}
&\text{Pr}_{\boldsymbol f}(Y=i|X=x) \\
\approx & A(q,p_i) \\
=& \frac{\exp(\mathbf{q}_q \mathbf{k}_{p_i}^{ T}/\sqrt{d})}{\sum_{j=1}^N \exp(\mathbf{q}_q \mathbf{k}_{j}^{T}/\sqrt{d})}.
\end{aligned}
\end{equation}
By setting $\mathbf{q}_q/\sqrt{d} = \hat{\mathbf{x}}$ and $\mathbf{k}_{p_i} - \mathbf{k}_{p_C} = \boldsymbol{\beta}_i$, we deduce:
\begin{equation}
\label{equ:icl}
\small 
\log\frac{\text{Pr}_{\boldsymbol f}(Y=i|X=x)}{\text{Pr}_{\boldsymbol f}(Y=C|X=x)}=\boldsymbol{\beta}_{i}^{T} \hat{\mathbf{x}}.
\end{equation}
This approximates a logistic regression model where:
\begin{equation}
\small 
\log\frac{\text{Pr}_{\boldsymbol f}(Y=i|X=x)}{\text{Pr}_{\boldsymbol f}(Y=C|X=x)} =\beta_0^i+ \boldsymbol{\beta}_{i}^{T} \mathbf{x}.
\end{equation}
In this equation, $\beta_0^i$ and $\boldsymbol{\beta}_{i}^{T}$ are parameters that can be learned, while $\mathbf{x}$ is the input feature.

Inspired by the similarity between ICL and logistic regression, we've incorporated a learnable $\beta_0^i$ into Eq.~(\ref{equ:icl}), which is equivalent to adjusting the attention weights $A(q,p_i)$:
\begin{equation}
    \hat{A}(q,p_i) = \exp(\beta_0^i)A(q,p_i)
\end{equation}
% \begin{multline}
% \small 
%      \boldsymbol f(x) =
%      ( \exp(\beta^1_{lh}) A_l^h\left(q, p_1\right), \\ \ldots,\exp(\beta^C_{lh})A_l^h\left(q, p_C\right)),
% \end{multline}
Each $\beta_0^i$ is a learnable parameter, set uniquely for different attention heads and layers. Refer to Appendix~\ref{appendix:anchor_re-weighting} for more details.

To train the re-weighting vector $\boldsymbol{\beta} = \left\{\beta_0^{i} \right\}$, we utilize an auxiliary training set $(\boldsymbol{X}_{train},\boldsymbol{Y}_{train})$. Here, we perform ICL with normal demonstrations and optimize $\boldsymbol{\beta}$ with respect to the classification loss $\mathcal L$ on $(\boldsymbol{X}_{train},\boldsymbol{Y}_{train})$:

\begin{equation}
    \boldsymbol{\beta}^\star   = \arg\min_{\boldsymbol{\beta}}\mathcal L(\boldsymbol{X}_{train},\boldsymbol{Y}_{train}).
\end{equation}

% This can be metaphorically equated to ``re-weighting the anchors'', hence we refer to it as \textbf{Anchor Re-weighting}.
% It can also be viewed as a modification of the demonstration contributions since demonstration information has been incorporated into the anchors as suggested by our prior analysis in \S~\ref{subsec:shallow_layer}.
% Besides, it can be seen as a special type of adapter, since it introduces a small number of parameters and keeps most parts of the model as original, but it is designed based on our anchor hypothesis and owns fewer parameters than normal adapters. 
This approach can be metaphorically described as "re-weighting the anchors," leading us to term it as \textbf{Anchor Re-weighting}. It can also be viewed as a modification of the demonstration contributions since demonstration information has been incorporated into the anchors as suggested by our prior analysis in \S~\ref{subsec:shallow_layer}. Additionally, it can be interpreted as a unique adapter variant, introducing minimal parameters while preserving most of the original model. However, it is specifically designed based on our anchor hypothesis and requires fewer parameters than traditional adapters.

\subsubsection{Experiments}
\label{para:re-weighting_experiments}

We choose one sample per class as normal demonstrations and choose four extra samples per class to form the auxiliary training set $(\boldsymbol{X}_{train},\boldsymbol{Y}_{train})$. The setup follows \S~\ref{para:experimental_settings_shallow}, with results averaged over five random seeds. Owing to computational constraints, we employ GPT2-XL for evaluation, excluding GPT-J. The parameters $\left\{\beta_0^{i} \right\}$ are trained using gradient descent. More details can be found in Appendix~\ref{appendix: anchor-re-weighting-settings}.
% \wang{appendix not done}

We compare \textbf{Anchoring Re-weighting} with two baselines: (1) Vanilla ICL with the same demonstration (1-shot per class) (2) Vanilla ICL, where the auxiliary training set of $\boldsymbol \beta$ is included as demonstrations (5-shot per class) for a fair comparison.

\subsubsection{Results}
\label{para:re-weighting_results}
% 加个结论

As Table \ref{tab:lda_1} shows, the proposed anchor re-weighting significantly enhances ICL performance, particularly on the SST-2 and EmoC datasets. Besides, adding more demonstrations for vanilla ICL may not bring a stable accuracy boost due to the potential noise introduced, as discussed in \citet{Zhao2021CalibrateBU}.
Different from vanilla ICL which utilizes the extra examples to form a demonstration, we train a re-weighting vector $\boldsymbol \beta$ to modulate label anchor contributions. This shortens the input context and thus brings (almost) no extra cost to the inference speed. The consistent improvements of our method suggest that the re-weighting mechanism could be a better alternative to utilize demonstration examples. Furthermore, it reiterates the crucial role that anchors play in ICL.

% \subsection{Input Compression for ICL}
\subsection{Anchor-Only Context Compression}
\label{subsec:compress}

We further explore a context compression technique that reduces the full demonstration to anchor hidden states for accelerating ICL inference.

\subsubsection{Method}
\label{para:compression_method}

In \S~\ref{subsec:deep_layer}, we find that the model output heavily relies on the label words, which collect information from the demonstrations. Given the auto-regressive nature of GPT-like models, where hidden states of tokens depend solely on preceding ones, label words' information aggregation process is independent of subsequent words. This allows for the calculation and caching of the label word hidden states $\boldsymbol H = \{\{\boldsymbol h_l^i\}_{i=1}^C\}_{l=1}^N$ ($\boldsymbol h_l^i$ is the $l$-th layer's hidden state of the $i$-th label word in the demonstration).
By concatenating $\boldsymbol h_l^1,...,\boldsymbol h_l^C$ at the front in each layer during inference, instead of using the full demonstration, we can speed up inference.
%\lei{symbol meaning such as T(x) V(y) needs to be specified here}

In our preliminary experiments, concatenating hidden states of label words alone was inadequate for completing the ICL task.\footnote{Omitting formatting significantly reduces accuracy, as the model will favor common tokens like ``the'' over label words, indicating confusion about the expected output type.} This might be due to the critical role of formatting information in helping the model to determine the output space at the target position,\footnote{Here, ``formatting'' refers to elements like ``Review:'' and ``Sentiment:'' in Figure~\ref{fig:icl_diagram}.} as highlighted in \citet{Min2022RethinkingTR}. As a solution, we amalgamate the hidden states of both the formatting and the label words, a method we've termed \textbf{Hidden$_{\textrm{anchor}}$}.

\subsubsection{Experiments}
\label{para:compress_results}
We follow the same experimental settings as \S~\ref{para:experimental_settings_shallow}.
We compare our {Hidden$_{\textrm{anchor}}$} input compression method with two equally efficient baselines.

\noindent\textbf{Text$_{\textrm{anchor}}$}: This method concatenates the formatting and label text with the input, as opposed to concatenating the hidden states at each layer.

\noindent\textbf{Hidden$_{\textrm{random}}$}: This approach concatenates the hidden states of formatting and randomly selected non-label words (equal in number to {Hidden$_\textrm{anchor}$}). 

\noindent\textbf{Hidden$_{\textrm{random-top}}$}: To establish a stronger baseline, we randomly select 20 sets of non-label words in Hidden$_{\textrm{random}}$ and report the one with the highest label loyalty. %Randomly selecting the same number of hidden states as \textbf{Hidden$_\textrm{anchor}$} from all hidden states of the demonstrations for concatenation.

The \textbf{Text$_{\textrm{anchor}}$} method is included to demonstrate that the effectiveness of {Hidden$_{\textrm{anchor}}$} is attributed to the aggregation of information in label words, rather than the mere text of label words. If we find that {Hidden$_{\textrm{anchor}}$} surpasses {Text$_{\textrm{anchor}}$} in performance, it solidifies the notion that the aggregated information within label words carries significant importance. The \textbf{Hidden$_{\textrm{random}}$} method is introduced to illustrate that anchor hidden states encapsulate most of the demonstration information among all hidden states.

We assess all compression methods using the label loyalty and word loyalty introduced in \S~\ref{subsec:shallow_layer}, in addition to classification accuracy.

\subsubsection{Results}
% 图LL, WL等的缩写尽量改全名

\begin{table}[t!]
\small 
  \centering
  \begin{tabular}{ @{} l|c@{\hspace{0.3cm}}c@{\hspace{0.3cm}}c@{}}
    \toprule
     Method & Label Loyalty & Word Loyalty & Acc. \\
    \midrule
     ICL~(GPT2-XL)& $100.00$ & $100.00$ & $51.90$ \\
     \midrule
Text$_{\textrm{anchor}}$ & $51.05$ & $36.65$ & $38.77$ \\
Hidden$_{\textrm{random}}$ & $48.96$ & $5.59$ & $39.96$ \\
Hidden$_{\textrm{random-top}}$ & $57.52$ & $4.49$ & $41.72$ \\
Hidden$_{\textrm{anchor}}$ & $\boldsymbol{79.47}$ & $\boldsymbol{62.17}$ & $\boldsymbol{45.04}$ \\
    \midrule
 ICL~(GPT-J) & $100.00$ & $100.00$ & $56.82$ \\
     \midrule
 
Text$_{\textrm{anchor}}$ & $53.45$ & $43.85$ & $40.83$ \\
Hidden$_{\textrm{random}}$ & $49.03$ & $2.16$ & $31.51$ \\
Hidden$_{\textrm{random-top}}$ & $71.10$ & $11.36$ & $52.34$ \\
Hidden$_{\textrm{anchor}}$ & $\boldsymbol{89.06}$ & $\boldsymbol{75.04}$ & $\boldsymbol{55.59}$ \\
    \bottomrule
  \end{tabular}
    \caption{Results of different compression methods on GPT2-XL and GPT-J (averaged over SST-2, TREC, AGNews, and EmoC). Acc. denotes accuracy. The best results are shown in bold. Our method achieves the best compression performance.}
     \label{tab:compress-gpt2}
% updated with new Hidden_random
\end{table}

% \begin{table}[t!]
%   \centering
%   \small

%   \begin{tabular}{@{\hspace{-0.1cm}} c|ccc}
%     \toprule
%      Method & Label Loyalty & Word Loyalty & Accuracy  \\

%     \bottomrule
%   \end{tabular}
%     \caption{Results of different compression methods on GPT-J. Our method achieves the best performance among compression methods.}
%   \label{tab:compress-gptj}
% \end{table}

We can see from Table~\ref{tab:compress-gpt2} that the proposed compression method \textbf{Hidden$_{\textrm{anchor}}$} achieves the best results among all three compression methods on all metrics and for both models. For example, with the GPT-J model, the compression method with anchor states only leads to a $1.5$ accuracy drop compared to the uncompressed situation, indicating that the compression introduces negligible information loss.
Further, we estimate the efficiency improvements over the original ICL.
As shown in Table~\ref{tab:compress_eff},
the speed-up ratio ranges from $1.1\times$ to $2.9\times$, as the efficiency gain is influenced by the length of the demonstrations. We refer readers to Appendix~\ref{appendix:L_demo} for a more elaborated analysis of the speed-up ratios.
Besides, we observe that the  acceleration effect is more pronounced in the GPT-J model compared to GPT2-XL, demonstrating its great potential to apply to larger language models.
% From the table, we can observe that the larger the ratio of total length to predicted text length, the more significant the efficiency improvement. This is especially evident in the TREC dataset, where the acceleration ratio exceeds 2.5$\times$. 

% (Appendix)

\begin{table}[t!]
  \centering
\small
  \begin{tabular}{c|cccc}
    \toprule
     Model & SST-2 & TREC & AGNews & EmoC \\
    \midrule
    GPT2-XL & $1.1\times$ & $1.5\times$ & $2.5\times$ & $1.4\times$ \\
    GPT-J & $1.5\times$ & $2.2\times$ & $2.9\times$ & $1.9\times$\\
    %$L_{\text{demo}}/L_{\mathbf x}$ & $18/19$ & $61/7$ & $151/37$ & $53/12$ \\
    \bottomrule
  \end{tabular}
    \caption{Acceleration ratios of the Hidden$_{\textrm{anchor}}$ method.}
  \label{tab:compress_eff}
\end{table}

\subsection{Anchor Distances for Error Diagnosis}
\label{subsec:error_analysis}

Lastly, we perform an error analysis for ICL by examining the distances between the key vectors in the attention module that correspond to the label words. 

\subsubsection{Method}
\label{para:dignoise_method}
Our previous analysis in \S~\ref{subsec:deep_layer} shows a strong correlation between the model output and $A(q,p_i)$, which is determined by $\mathbf{q}_q\mathbf{k}_{p_i}^T$ as per Eq.~\ref{equ:icl}. Should the key vectors $\mathbf{k}$ for label words $p_i$ and $p_k$ be similar, $A(q,p_i)$ and $A(q,p_k)$ will also likely be similar, leading to potential label confusion. Furthermore, considering the distribution of query vectors $\mathbf{q}_q$, we employ a PCA-like method to extract the components of the key vectors along the directions with significant variations in $\mathbf{q}_q$, denoted as $\hat{\mathbf{k}}$ (see Appendix~\ref{appendix:hat_k} for details). We anticipate that the distances between these $\hat{\mathbf{k}}$s can correspond to the category confusion of the model, thus revealing one possible origin of ICL errors. Here, we normalize the distances to a scale of 0-1, with 0 indicating the highest degree of category confusion:
\begin{equation}
    \text {Confusion}_{ij}^{\text{pred}} = \frac{ \|\hat{\mathbf{k_{p_i}}} - \hat{\mathbf{k_{p_j}}}\|}{\max_{s\not=t} \|\hat{\mathbf{k_{p_s}}} - \hat{\mathbf{k_{p_t}}}\|},
\end{equation}

\subsubsection{Experiments}
We utilize the GPT2-XL model and TREC dataset, as the model displays varying confusion levels between categories on this dataset. We use all 500 samples of the TREC test set and use 1 demonstration per class for convenience of analysis.

We calculate the actual model confusion score, $\text{Confusion}_{ij}$, between category $i$ and category $k$ using the AUC-ROC metric (detailed in Appendix~\ref{appendix:confusion}). We then compare the predicted confusion score, $\text{Confusion}_{ij}^{\text{pred}}$, and the actual confusion score, $\text{Confusion}_{ij}$, via heatmaps.
%\lei{add dataset details here, 1000 samples as well?}

% \subsubsection{Experiments}

 % and $\text{Confusion}_{ij}$ for comparison.

\subsubsection{Results}
\label{para:diagnoise_results}

\begin{figure}[t!]
\centering
\subfloat[Confusion matrix of $\text {Confusion}_{ij}^{\text{pred}}$.]{
    \label{sfig:cbp}
    \includegraphics[height=5.2cm]{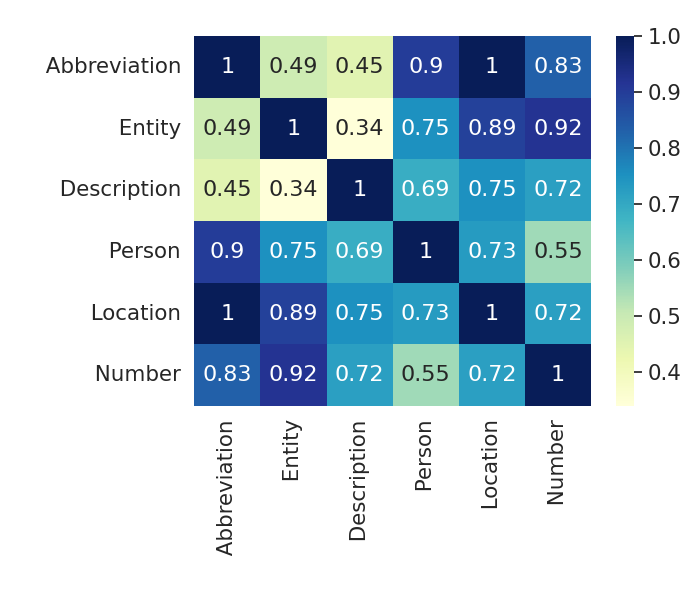}}
    \hspace{2em}
    % \hfill
\subfloat[Confusion matrix of $\text {Confusion}_{ij}$.]{
    \label{sfig:cb}
    \includegraphics[height=5.2cm]{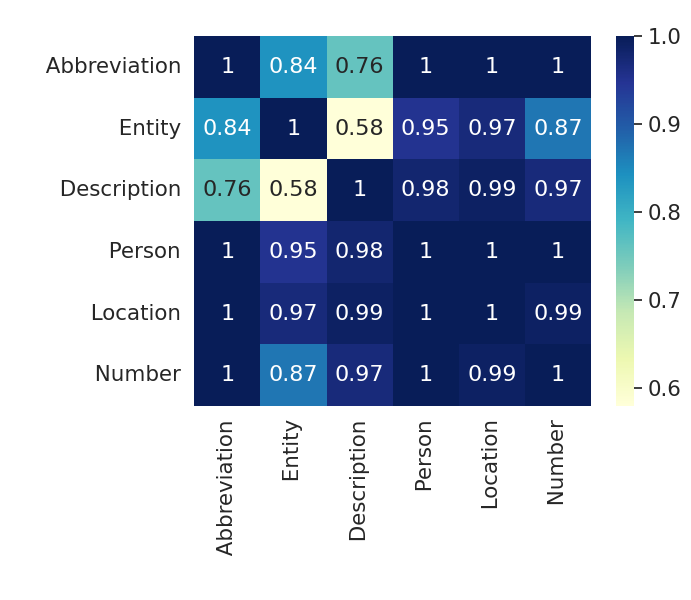}}
    
  \caption{Predicted and real confusion matrix on TREC. We set undefined diagonals to $1$ for better visualization. The heatmaps display similarity in confusing category pairs, particularly in lighter-colored blocks.}
  \label{fig:conf}

% attention_attr_ana.ipynb
\end{figure}
% The heatmap for $\text{Confusion}_{ij}^{\text{pred}}$
 % and $\text{Confusion}_{ij}$ is 
 Figure~\ref{fig:conf} shows that the proposed approximation metric, $\text{Confusion}_{ij}^{\text{pred}}$, can identify the most confusing case (Description-Entity) and performs reasonably well for highly confusing categories (Entity-Abbreviation, Description-Abbreviation). 
 This high correlation indicates that ICL makes errors in categories with similar label anchors.
 Overall, this result demonstrates that our anchor-based analysis framework could serve as an interpretation tool for better understanding ICL's errors.
 % For other cases, where the confusion is not severe, the alignment is relatively poorer.

 % Here, lower values (lighter colors) indicate higher confusion levels.
% It can be observed that 

\section{Related Work}
% 先分一下类,总结一下,我们的方法和后面的比较像
% With the pursuit of a better understanding of the working mechanism of in-context learning, 
% There have been many analytical works, which can be roughly categorized into two streams.

The existing literature on in-context learning analysis can be broadly divided into two streams, each focusing on different aspects.
The first stream explores the influencing factors of ICL based on input perturbation, such as the order~\citep{Min2022RethinkingTR}, the formatting~\citep{Kim2022GroundTruthLM,Wei2022ChainOT}, and the selection of the demonstration~\citep{Liu2021WhatMG}.
Designing proper demonstration construction strategies~\citep{ye2023ceil,udr} and calibration techniques~\citep{Zhao2021CalibrateBU,Min2021NoisyCL} could bring clear boosts to the ICL performance. 
The second stream investigates the inner working mechanism of ICL through different conceptual lenses, such as making an analogy of ICL to gradient descent~\citep{Oswald2022TransformersLI, dai2022can} and viewing the process of ICL as a Bayesian inference~\citep{Xie2021AnEO}.

In this paper, we provide a novel perspective by examining the information flow in language models to gain an understanding of ICL. Our approach offers new insights and demonstrates the potential for leveraging this understanding to improve the effectiveness, efficiency, and interpretability of ICL.

\section{Conclusion}
In this paper, we propose a hypothesis that label words serve as anchors in in-context learning for aggregating and distributing the task-relevant information flow.
Experimental results with attention manipulation and analysis of predictions correlation consolidate the hypothesis holds well in GPT2-XL and GPT-J models.
Inspired by the new understanding perspective, we propose three 
practical applications.
First, an anchor re-weighting method is proposed to improve ICL accuracy. Second, we explore a demonstration compression technique to accelerate ICL inference. Lastly, we showcase an analysis framework to diagnose ICL errors on a real-world dataset.
These promising applications again verify the hypothesis and open up new directions for future investigations on ICL. 

\section*{Limitations}
Our study, while providing valuable insights into in-context learning~(ICL), has several limitations. Firstly, our research scope was limited to classification tasks and did not delve into the realm of generative tasks. Additionally, our hypothesis was only examined within conventional ICL paradigms, leaving other ICL paradigms such as the chain of thought prompting~(CoT)~\cite{Wei2022ChainOT} unexplored. Secondly, due to hardware constraints, we mainly investigated models up to a scale of 6 billion parameters. Further research that replicates our study using larger-scale models would be beneficial in corroborating our findings and refining the hypotheses set forth in our investigation.

\section*{Acknowledgement}
We thank all reviewers for their thoughtful and insightful suggestions. This work is supported in part by a Tencent Research Grant and National Natural Science Foundation of China (No. 62176002). Xu Sun is the corresponding author.

\bibliography{custom}

\begin{thebibliography}{28}
\expandafter\ifx\csname natexlab\endcsname\relax\def\natexlab#1{#1}\fi

\bibitem[{Aky{\"u}rek et~al.(2022)Aky{\"u}rek, Schuurmans, Andreas, Ma, and
  Zhou}]{Akyrek2022WhatLA}
Ekin Aky{\"u}rek, Dale Schuurmans, Jacob Andreas, Tengyu Ma, and Denny Zhou.
  2022.
\newblock \href {https://arxiv.org/abs/2211.15661} {What learning algorithm is
  in-context learning? investigations with linear models}.
\newblock \emph{ArXiv preprint}, abs/2211.15661.

\bibitem[{Brown et~al.(2020)Brown, Mann, Ryder, Subbiah, Kaplan, Dhariwal,
  Neelakantan, Shyam, Sastry, Askell, Agarwal, Herbert{-}Voss, Krueger,
  Henighan, Child, Ramesh, Ziegler, Wu, Winter, Hesse, Chen, Sigler, Litwin,
  Gray, Chess, Clark, Berner, McCandlish, Radford, Sutskever, and
  Amodei}]{Brown2020LanguageMA}
Tom~B. Brown, Benjamin Mann, Nick Ryder, Melanie Subbiah, Jared Kaplan,
  Prafulla Dhariwal, Arvind Neelakantan, Pranav Shyam, Girish Sastry, Amanda
  Askell, Sandhini Agarwal, Ariel Herbert{-}Voss, Gretchen Krueger, Tom
  Henighan, Rewon Child, Aditya Ramesh, Daniel~M. Ziegler, Jeffrey Wu, Clemens
  Winter, Christopher Hesse, Mark Chen, Eric Sigler, Mateusz Litwin, Scott
  Gray, Benjamin Chess, Jack Clark, Christopher Berner, Sam McCandlish, Alec
  Radford, Ilya Sutskever, and Dario Amodei. 2020.
\newblock \href
  {https://proceedings.neurips.cc/paper/2020/hash/1457c0d6bfcb4967418bfb8ac142f64a-Abstract.html}
  {Language models are few-shot learners}.
\newblock In \emph{Advances in Neural Information Processing Systems 33: Annual
  Conference on Neural Information Processing Systems 2020, NeurIPS 2020,
  December 6-12, 2020, virtual}.

\bibitem[{Chatterjee et~al.(2019)Chatterjee, Narahari, Joshi, and
  Agrawal}]{chatterjee-etal-2019-semeval}
Ankush Chatterjee, Kedhar~Nath Narahari, Meghana Joshi, and Puneet Agrawal.
  2019.
\newblock \href {https://doi.org/10.18653/v1/S19-2005} {{S}em{E}val-2019 task
  3: {E}mo{C}ontext contextual emotion detection in text}.
\newblock In \emph{Proceedings of the 13th International Workshop on Semantic
  Evaluation}, pages 39--48, Minneapolis, Minnesota, USA. Association for
  Computational Linguistics.

\bibitem[{Dai et~al.(2022)Dai, Sun, Dong, Hao, Sui, and Wei}]{dai2022can}
Damai Dai, Yutao Sun, Li~Dong, Yaru Hao, Zhifang Sui, and Furu Wei. 2022.
\newblock \href {https://arxiv.org/abs/2212.10559} {Why can gpt learn
  in-context? language models secretly perform gradient descent as meta
  optimizers}.
\newblock \emph{ArXiv preprint}, abs/2212.10559.

\bibitem[{Dong et~al.(2023)Dong, Li, Dai, Zheng, Wu, Chang, Sun, Xu, and
  Sui}]{icl_survey}
Qingxiu Dong, Lei Li, Damai Dai, Ce~Zheng, Zhiyong Wu, Baobao Chang, Xu~Sun,
  Jingjing Xu, and Zhifang Sui. 2023.
\newblock \href {https://arxiv.org/abs/2301.00234} {A survey for in-context
  learning}.
\newblock \emph{ArXiv preprint}, abs/2301.00234.

\bibitem[{Hovy et~al.(2001)Hovy, Gerber, Hermjakob, Lin, and
  Ravichandran}]{hovy-etal-2001-toward}
Eduard Hovy, Laurie Gerber, Ulf Hermjakob, Chin-Yew Lin, and Deepak
  Ravichandran. 2001.
\newblock \href {https://aclanthology.org/H01-1069} {Toward semantics-based
  answer pinpointing}.
\newblock In \emph{Proceedings of the First International Conference on Human
  Language Technology Research}.

\bibitem[{Kingma and Ba(2015)}]{Kingma2014AdamAM}
Diederik~P. Kingma and Jimmy Ba. 2015.
\newblock \href {http://arxiv.org/abs/1412.6980} {Adam: {A} method for
  stochastic optimization}.
\newblock In \emph{3rd International Conference on Learning Representations,
  {ICLR} 2015, San Diego, CA, USA, May 7-9, 2015, Conference Track
  Proceedings}.

\bibitem[{Kobayashi et~al.(2020)Kobayashi, Kuribayashi, Yokoi, and
  Inui}]{Kobayashi2020AttentionIN}
Goro Kobayashi, Tatsuki Kuribayashi, Sho Yokoi, and Kentaro Inui. 2020.
\newblock \href {https://doi.org/10.18653/v1/2020.emnlp-main.574} {Attention is
  not only a weight: Analyzing transformers with vector norms}.
\newblock In \emph{Proceedings of the 2020 Conference on Empirical Methods in
  Natural Language Processing (EMNLP)}, pages 7057--7075, Online. Association
  for Computational Linguistics.

\bibitem[{Li et~al.(2023{\natexlab{a}})Li, Lv, Yan, Lin, Zhu, Ni, Xie, Wang,
  and Qiu}]{udr}
Xiaonan Li, Kai Lv, Hang Yan, Tianyang Lin, Wei Zhu, Yuan Ni, Guotong Xie,
  Xiaoling Wang, and Xipeng Qiu. 2023{\natexlab{a}}.
\newblock \href {https://arxiv.org/abs/2305.04320} {Unified demonstration
  retriever for in-context learning}.
\newblock \emph{ArXiv preprint}, abs/2305.04320.

\bibitem[{Li and Roth(2002)}]{li-roth-2002-learning}
Xin Li and Dan Roth. 2002.
\newblock \href {https://aclanthology.org/C02-1150} {Learning question
  classifiers}.
\newblock In \emph{{COLING} 2002: The 19th International Conference on
  Computational Linguistics}.

\bibitem[{Li et~al.(2023{\natexlab{b}})Li, Ildiz, Papailiopoulos, and
  Oymak}]{Li2023TransformersAA}
Yingcong Li, Muhammed~Emrullah Ildiz, Dimitris Papailiopoulos, and Samet Oymak.
  2023{\natexlab{b}}.
\newblock Transformers as algorithms: Generalization and stability in
  in-context learning.

\bibitem[{Liu et~al.(2022)Liu, Shen, Zhang, Dolan, Carin, and
  Chen}]{Liu2021WhatMG}
Jiachang Liu, Dinghan Shen, Yizhe Zhang, Bill Dolan, Lawrence Carin, and Weizhu
  Chen. 2022.
\newblock \href {https://doi.org/10.18653/v1/2022.deelio-1.10} {What makes good
  in-context examples for {GPT}-3?}
\newblock In \emph{Proceedings of Deep Learning Inside Out (DeeLIO 2022): The
  3rd Workshop on Knowledge Extraction and Integration for Deep Learning
  Architectures}, pages 100--114, Dublin, Ireland and Online. Association for
  Computational Linguistics.

\bibitem[{Michel et~al.(2019)Michel, Levy, and Neubig}]{Michel2019AreSH}
Paul Michel, Omer Levy, and Graham Neubig. 2019.
\newblock \href
  {https://proceedings.neurips.cc/paper/2019/hash/2c601ad9d2ff9bc8b282670cdd54f69f-Abstract.html}
  {Are sixteen heads really better than one?}
\newblock In \emph{Advances in Neural Information Processing Systems 32: Annual
  Conference on Neural Information Processing Systems 2019, NeurIPS 2019,
  December 8-14, 2019, Vancouver, BC, Canada}, pages 14014--14024.

\bibitem[{Min et~al.(2022{\natexlab{a}})Min, Lewis, Hajishirzi, and
  Zettlemoyer}]{Min2021NoisyCL}
Sewon Min, Mike Lewis, Hannaneh Hajishirzi, and Luke Zettlemoyer.
  2022{\natexlab{a}}.
\newblock \href {https://doi.org/10.18653/v1/2022.acl-long.365} {Noisy channel
  language model prompting for few-shot text classification}.
\newblock In \emph{Proceedings of the 60th Annual Meeting of the Association
  for Computational Linguistics (Volume 1: Long Papers)}, pages 5316--5330,
  Dublin, Ireland. Association for Computational Linguistics.

\bibitem[{Min et~al.(2022{\natexlab{b}})Min, Lyu, Holtzman, Artetxe, Lewis,
  Hajishirzi, and Zettlemoyer}]{Min2022RethinkingTR}
Sewon Min, Xinxi Lyu, Ari Holtzman, Mikel Artetxe, Mike Lewis, Hannaneh
  Hajishirzi, and Luke Zettlemoyer. 2022{\natexlab{b}}.
\newblock \href {https://aclanthology.org/2022.emnlp-main.759} {Rethinking the
  role of demonstrations: What makes in-context learning work?}
\newblock In \emph{Proceedings of the 2022 Conference on Empirical Methods in
  Natural Language Processing}, pages 11048--11064, Abu Dhabi, United Arab
  Emirates. Association for Computational Linguistics.

\bibitem[{Radford et~al.(2019)Radford, Wu, Child, Luan, Amodei, Sutskever
  et~al.}]{Radford2019LanguageMA}
Alec Radford, Jeffrey Wu, Rewon Child, David Luan, Dario Amodei, Ilya
  Sutskever, et~al. 2019.
\newblock Language models are unsupervised multitask learners.
\newblock \emph{OpenAI blog}, 1(8):9.

\bibitem[{Simonyan et~al.(2013)Simonyan, Vedaldi, and
  Zisserman}]{Simonyan2013DeepIC}
Karen Simonyan, Andrea Vedaldi, and Andrew Zisserman. 2013.
\newblock Deep inside convolutional networks: Visualising image classification
  models and saliency maps.
\newblock \emph{CoRR}, abs/1312.6034.

\bibitem[{Socher et~al.(2013)Socher, Perelygin, Wu, Chuang, Manning, Ng, and
  Potts}]{socher-etal-2013-recursive}
Richard Socher, Alex Perelygin, Jean Wu, Jason Chuang, Christopher~D. Manning,
  Andrew Ng, and Christopher Potts. 2013.
\newblock \href {https://aclanthology.org/D13-1170} {Recursive deep models for
  semantic compositionality over a sentiment treebank}.
\newblock In \emph{Proceedings of the 2013 Conference on Empirical Methods in
  Natural Language Processing}, pages 1631--1642, Seattle, Washington, USA.
  Association for Computational Linguistics.

\bibitem[{Touvron et~al.(2023)Touvron, Lavril, Izacard, Martinet, Lachaux,
  Lacroix, Rozi{\`e}re, Goyal, Hambro, Azhar, Rodriguez, Joulin, Grave, and
  Lample}]{Touvron2023LLaMAOA}
Hugo Touvron, Thibaut Lavril, Gautier Izacard, Xavier Martinet, Marie-Anne
  Lachaux, Timoth{\'e}e Lacroix, Baptiste Rozi{\`e}re, Naman Goyal, Eric
  Hambro, Faisal Azhar, Aurelien Rodriguez, Armand Joulin, Edouard Grave, and
  Guillaume Lample. 2023.
\newblock \href {https://api.semanticscholar.org/CorpusID:257219404} {Llama:
  Open and efficient foundation language models}.
\newblock \emph{ArXiv}, abs/2302.13971.

\bibitem[{von Oswald et~al.(2022)von Oswald, Niklasson, Randazzo, Sacramento,
  Mordvintsev, Zhmoginov, and Vladymyrov}]{Oswald2022TransformersLI}
Johannes von Oswald, Eyvind Niklasson, E.~Randazzo, Jo{\~a}o Sacramento,
  Alexander Mordvintsev, Andrey Zhmoginov, and Max Vladymyrov. 2022.
\newblock \href {https://arxiv.org/abs/2212.07677} {Transformers learn
  in-context by gradient descent}.
\newblock \emph{ArXiv preprint}, abs/2212.07677.

\bibitem[{Wang and Komatsuzaki(2021)}]{gpt-j}
Ben Wang and Aran Komatsuzaki. 2021.
\newblock {GPT-J-6B: A 6 Billion Parameter Autoregressive Language Model}.
\newblock \url{https://github.com/kingoflolz/mesh-transformer-jax}.

\bibitem[{Wei et~al.(2022)Wei, Wang, Schuurmans, Bosma, hsin Chi, Xia, Le, and
  Zhou}]{Wei2022ChainOT}
Jason Wei, Xuezhi Wang, Dale Schuurmans, Maarten Bosma, Ed~Huai hsin Chi,
  F.~Xia, Quoc Le, and Denny Zhou. 2022.
\newblock \href {https://arxiv.org/abs/2201.11903} {Chain of thought prompting
  elicits reasoning in large language models}.
\newblock \emph{ArXiv preprint}, abs/2201.11903.

\bibitem[{Wei et~al.(2023)Wei, Wei, Tay, Tran, Webson, Lu, Chen, Liu, Huang,
  Zhou, and Ma}]{Wei2023LargerLM}
Jerry~W. Wei, Jason Wei, Yi~Tay, Dustin Tran, Albert Webson, Yifeng Lu, Xinyun
  Chen, Hanxiao Liu, Da~Huang, Denny Zhou, and Tengyu Ma. 2023.
\newblock \href {https://api.semanticscholar.org/CorpusID:257378479} {Larger
  language models do in-context learning differently}.
\newblock \emph{ArXiv}, abs/2303.03846.

\bibitem[{Xie et~al.(2022)Xie, Raghunathan, Liang, and Ma}]{Xie2021AnEO}
Sang~Michael Xie, Aditi Raghunathan, Percy Liang, and Tengyu Ma. 2022.
\newblock \href {https://openreview.net/forum?id=RdJVFCHjUMI} {An explanation
  of in-context learning as implicit bayesian inference}.
\newblock In \emph{The Tenth International Conference on Learning
  Representations, {ICLR} 2022, Virtual Event, April 25-29, 2022}.
  OpenReview.net.

\bibitem[{Ye et~al.(2023)Ye, Wu, Feng, Yu, and Kong}]{ye2023ceil}
Jiacheng Ye, Zhiyong Wu, Jiangtao Feng, Tao Yu, and Lingpeng Kong. 2023.
\newblock \href {https://arxiv.org/abs/2302.05698} {Compositional exemplars for
  in-context learning}.
\newblock \emph{ArXiv preprint}, abs/2302.05698.

\bibitem[{Yoo et~al.(2022)Yoo, Kim, Kim, Cho, Jo, Lee, Lee, and
  Kim}]{Kim2022GroundTruthLM}
Kang~Min Yoo, Junyeob Kim, Hyuhng~Joon Kim, Hyunsoo Cho, Hwiyeol Jo, Sang-Woo
  Lee, Sang-goo Lee, and Taeuk Kim. 2022.
\newblock \href {https://aclanthology.org/2022.emnlp-main.155} {Ground-truth
  labels matter: A deeper look into input-label demonstrations}.
\newblock In \emph{Proceedings of the 2022 Conference on Empirical Methods in
  Natural Language Processing}, pages 2422--2437, Abu Dhabi, United Arab
  Emirates. Association for Computational Linguistics.

\bibitem[{Zhang et~al.(2015)Zhang, Zhao, and LeCun}]{Zhang2015CharacterlevelCN}
Xiang Zhang, Junbo~Jake Zhao, and Yann LeCun. 2015.
\newblock \href
  {https://proceedings.neurips.cc/paper/2015/hash/250cf8b51c773f3f8dc8b4be867a9a02-Abstract.html}
  {Character-level convolutional networks for text classification}.
\newblock In \emph{Advances in Neural Information Processing Systems 28: Annual
  Conference on Neural Information Processing Systems 2015, December 7-12,
  2015, Montreal, Quebec, Canada}, pages 649--657.

\bibitem[{Zhao et~al.(2021)Zhao, Wallace, Feng, Klein, and
  Singh}]{Zhao2021CalibrateBU}
Zihao Zhao, Eric Wallace, Shi Feng, Dan Klein, and Sameer Singh. 2021.
\newblock \href {http://proceedings.mlr.press/v139/zhao21c.html} {Calibrate
  before use: Improving few-shot performance of language models}.
\newblock In \emph{Proceedings of the 38th International Conference on Machine
  Learning, {ICML} 2021, 18-24 July 2021, Virtual Event}, volume 139 of
  \emph{Proceedings of Machine Learning Research}, pages 12697--12706. {PMLR}.

\end{thebibliography}
\bibliographystyle{acl_natbib}

\appendix
\section*{Appendix}
% \section{Example Appendix}
% \label{sec:appendix}
\section{Experimental Settings}
\label{apx:dataset}
% \subsection{Models}
For models, we use GPT2-XL~(1.5B)~\cite{Radford2019LanguageMA} and GPT-J~(6B)~\cite{gpt-j} in this paper.

For datasets, we use a sentiment analysis task, Stanford Sentiment Treebank Binary~(SST-2)~\cite{socher-etal-2013-recursive}, a question type classification task, Text REtrieval Conference Question Classification~(TREC)~\cite{li-roth-2002-learning,hovy-etal-2001-toward}, a topic classification task, AG's news topic classification dataset~(AGNews)~\cite{Zhang2015CharacterlevelCN}, and an emotion classification task, EmoContext~(EmoC)~\cite{chatterjee-etal-2019-semeval}. The ICL templates of these tasks are shown in Table~\ref{tab:demo}.

\begin{table}[t!]
\small
  \centering
  \caption{Demonstration templates and label words. Here <S1> represents the demonstration, <S> represents the input to be predicted, and <L> represents the label word corresponding to the demonstration. To save space, we only show one demonstration for each task.}
  % {'text': 'Israel Destroys Refugee Homes, Kills One GAZA CITY, Gaza Strip - A day after a mortar round killed an Israeli-American woman in a nearby settlement, the Israeli army charged into a Palestinian refugee camp Saturday, killing one person and tearing down 35 homes, witnesses and a U.N. aid official said...',
 % 'label': 0}
  \label{tab:demo}
  % \begin{minipage}[t]{0.9\linewidth} %
  \begin{tabular}{clc}
    \toprule
    Task & Template & Label Words \\
    \midrule
    SST-2 & Review: <S1> & Positive, Negative \\
    & Sentiment: <L> & \\
    & Review: <S> & \\
    & Sentiment: & \\
    \midrule 
    TREC & Question: <S1> &  Abbreviation, Entity\\
    & Answer Type: <L> & Description, Person \\
    & Question: <S> & Location, Number \\
    & Answer Type: & \\
    \midrule
    AGNews & Article: <S1> &  World, Sports \\
    & Answer: <L> & Business, Technology \\
    & Article: <S> & \\
    & Answer: & \\
    \midrule
    EmoC & Dialogue: <S1> & Others, Happy\\
    & Emotion: <L> & Sad, Angry \\
    & Dialogue: <S> & \\
    & Emotion: &  \\
    \bottomrule
  \end{tabular}
  % \end{minipage}
\end{table}

\section{Results of $S_{wp}$, $S_{pq}$, and $S_{ww}$ on TREC and EmoC}
\label{sec:appdendix_trec_emoc}
% Relative sizes of $S_{wp}$, $S_{pq}$, and $S_{ww}$ of different layers on TREC and EmoC are shown in Figure~\ref{fig:attn_attr_2}. The results are similar that on SST-2 and AGNews. In shallow layers, $S_{pq}$, the significance of the information flow from label words to targeted positions, is low, while $S_{wp}$, the information flow from the text part to label words is high; (2) in deep layers, $S_{pq}$, the importance of information flow from label words to the targeted position becomes the dominant one. Notably, $S_{pq}$ and $S_{wp}$ usually surpass $S_{ww}$, suggesting that interactions involving label words outweigh others.

\begin{figure}[t!]
\centering
% \subfloat[Results on the SST-2 dataset]{
% \label{sfig:attn_attr_sst2}
% \includegraphics[height=5cm]{photos/attn_attr_sst2_1.pdf}}
% % \hspace{2em}
% \hfill
% \subfloat[Results on the TREC dataset]{
% \label{sfig:attn_attr_trec}
% \includegraphics[height=5cm]{photos/attn_attr_trec.pdf}}

% % \hfill

\subfloat[Results on the TREC dataset]{
\label{sfig:attn_attr_trec}
\includegraphics[height=5cm]{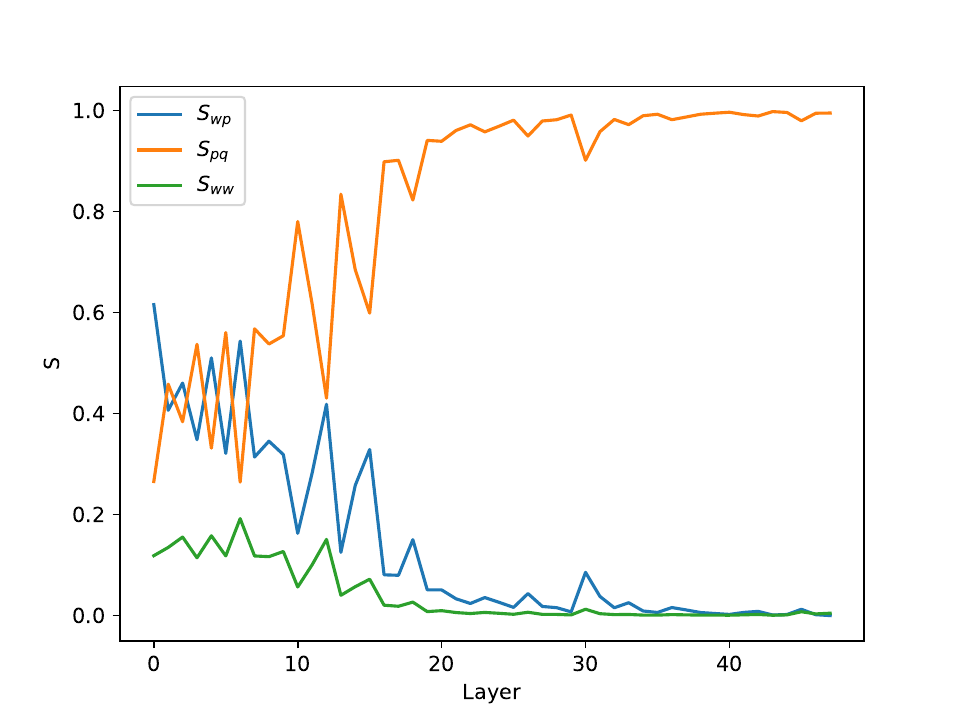}}
% \hspace{2em}
\hfill
\subfloat[Results on the EmoC dataset]{
\label{sfig:attn_attr_emoc}
\includegraphics[height=5cm]{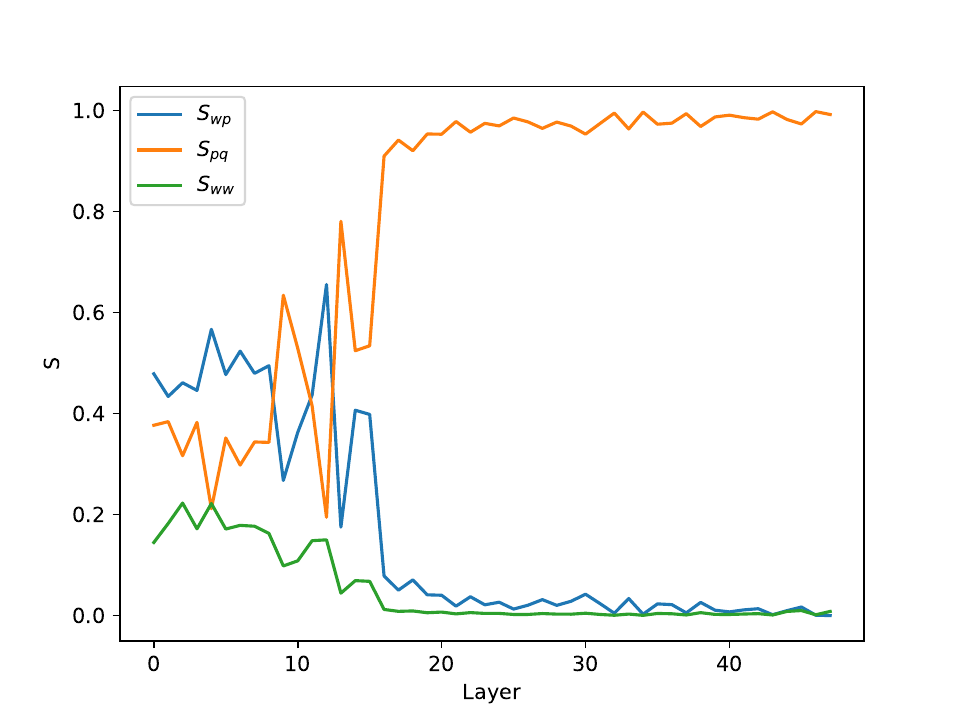}}

\caption{Relative size of $S_{wp}$, $S_{pq}$, and $S_{ww}$ on TREC and EmoC, which is similar to that on SST-2 and AGNews.}
\label{fig:attn_attr_2}
\end{figure}

Figure~\ref{fig:attn_attr_2} illustrates the relative sizes of $S_{wp}$, $S_{pq}$, and $S_{ww}$ on TREC and EmoC, mirroring results on SST-2 and AGNews. In shallow layers, $S_{wp}$~(the information flow from the text part to label words) is prominent, while $S_{pq}$~(the information flow from label words to targeted positions) is less significant. However, in deeper layers, $S_{pq}$ dominates. Importantly, $S_{wp}$ and $S_{pq}$ generally exceed $S_{ww}$, indicating that interactions involving label words are predominant.

\section{Reason for Using Word Loyalty Besides Label Loyalty}
\label{app:sec_wl_ll}

\begin{table*}[ht]
\centering
% {'text': 'Israel Destroys Refugee Homes, Kills One GAZA CITY, Gaza Strip - A day after a mortar round killed an Israeli-American woman in a nearby settlement, the Israeli army charged into a Palestinian refugee camp Saturday, killing one person and tearing down 35 homes, witnesses and a U.N. aid official said...',
% 'label': 0}
\begin{tabular}{ccc}
\toprule
Isolation Layer & Output Label & $V_5$ (sorted by probability)\\
\midrule
First 5 layers & World & ``\textbackslash n'', `` The'', `` Google'',``<|endoftext|>'', `` A'' \\
No isolation & World & `` World'', `` Technology'', `` Politics'', `` Israel'', `` Human''\\
\bottomrule
\end{tabular}
\caption{Results on a test sample with the label ``World'' from AGNews.}
\label{tab:agg-case}
\end{table*}

Label loyalty alone may not capture changes in the probability distribution of non-label words or the relative ratio of the probability of the label words within the entire vocabulary. Word loyalty helps address this limitation, which is shown in Table~\ref{tab:agg-case}.

\section{Isolating Different Numbers of Layers}
\label{appendix:isolating_different_num}

\begin{figure}[t!]
\centering
\subfloat[Effect of different numbers of isolated layers on GPT2-XL]{
\label{fig:aggregation_loyalty_gpt2-xl}
\includegraphics[height=5cm]{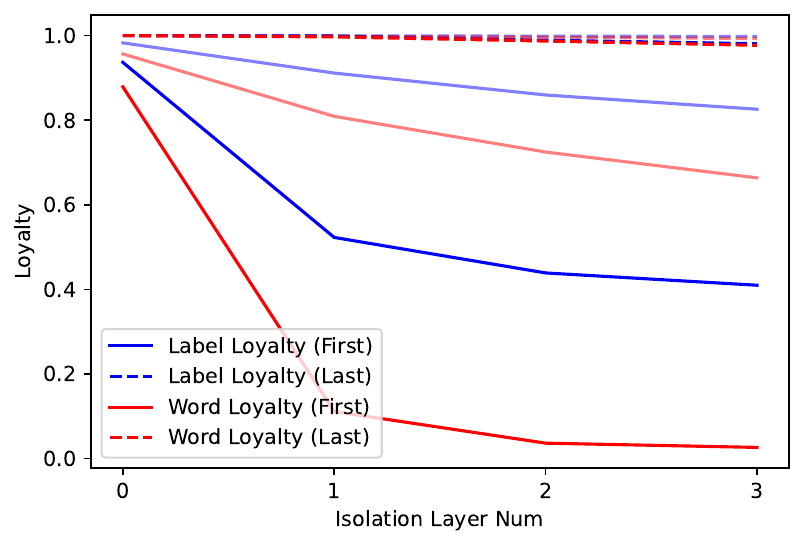}}
% \hspace{2em}
\hfill
\subfloat[Effect of different numbers of isolated layers on GPT-J]{
\label{fig:aggregation_loyalty_gpt-j}
\includegraphics[height=5cm]{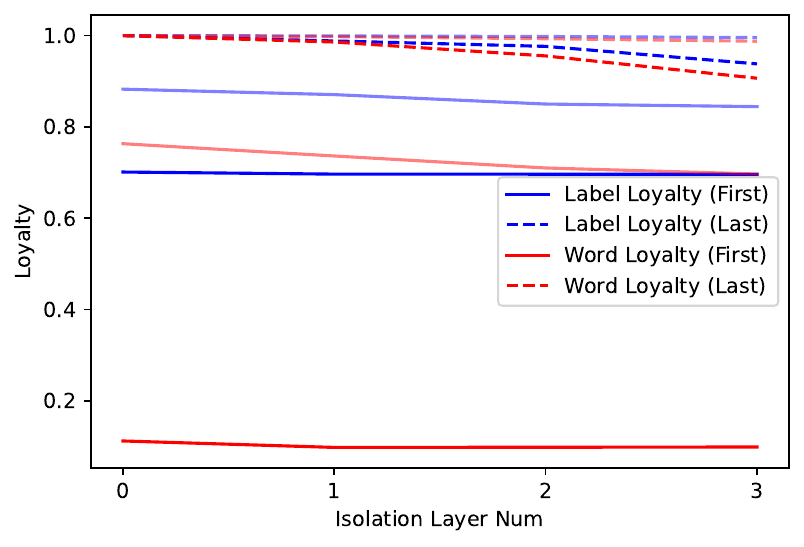}}
\caption{The chart demonstrates variations in label loyalty and word loyalty, dependent on whether label or non-label words are isolated in various layers. 'First' refers to the first several layers, while 'Last' to the last ones. Deep-colored lines represent label word isolation, whereas light colors denote non-label words. Remarkably, isolating label words in the shallow layers significantly influences the outcome, regardless of whether this is compared to isolation in deep layers or to non-label word isolation.}
\label{fig:aggregation_loyalty}
\end{figure}

We study the impact of the numbers of isolated layers, as shown in Figures \ref{fig:aggregation_loyalty_gpt2-xl} and \ref{fig:aggregation_loyalty_gpt-j}. It can be found that isolating shallow layers cause a significant impact, isolating deep layers has a negligible impact on the model, even when the number of isolation layers increases. This further illustrates the important role of information aggregation via label words in the shallow layers.

% 图放附录

\section{Details for the Calculation of $\text{AUCROC}_l$}
\label{appendix:calculation_aucroc}
Suppose the positions of the label words in the input $x$ are $p_1,...,p_C$ (without loss of generality, we suppose $p_i$ corresponds to the $i$th class), the targeted position is $q$, the sum of the attention matrices of all attention heads at the $l$ layer is $A_l$. We postulate that there's a strong correlation between the attention distributions on the label words of the target position $(A_l(q,p_1),..., A_l(q,p_C))$ and
the model's final prediction. We use the AUC-ROC score to quantify this correlation. We regard $(A_l(q,p_1),..., A_l(q,p_C))$ as a classifier's prediction for the model output label (that is, $A_l(q,p_i)$ is equivalent to the probability of model outputting label $i$), and compute the AUC-ROC value of this prediction relative to the actual model output. We denote this as $\text{AUCROC}_l$. For the case with more demonstrations (Appendix~\ref{appendix:demonstration_2}), we simply sum up all $A_l(q,p)$ of the same class.%The AUC-ROC value represents the area under the ROC curve of the classifier. 

\section{Additional Experimental Results}

\subsection{Results with More Demonstrations}
\label{appendix:demonstration_2}
\begin{figure}

\subfloat[Results on the SST-2 dataset]{
\label{sfig:attn_attr_sst2_more_demo}
\includegraphics[height=5cm]{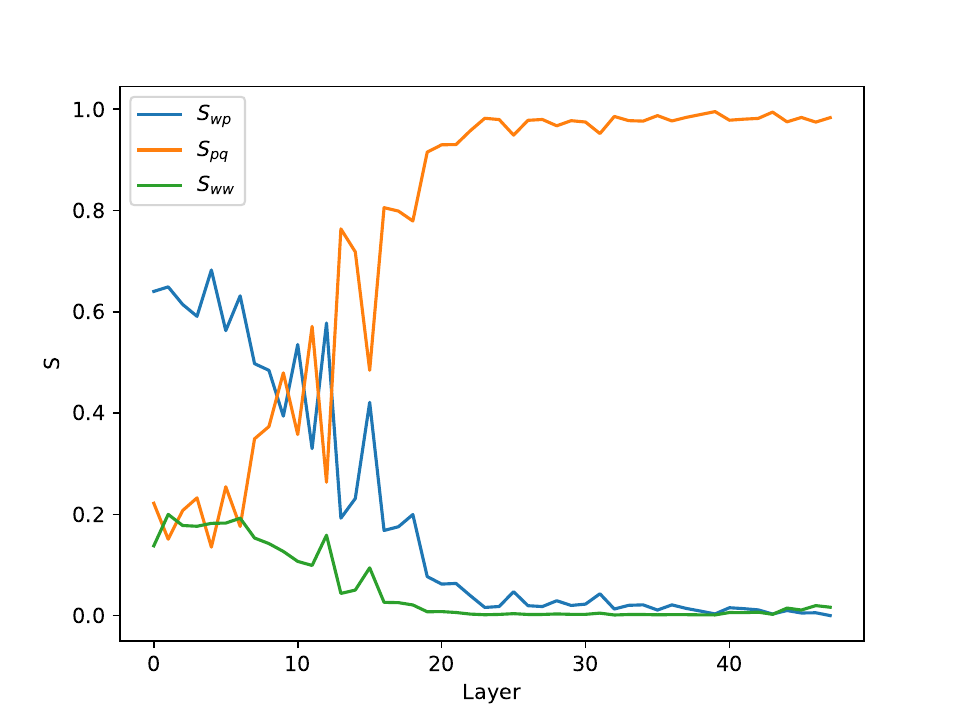}}
\hfill
\subfloat[Results on the TREC dataset]{
\label{sfig:attn_attr_trec_more_demo}
\includegraphics[height=5cm]{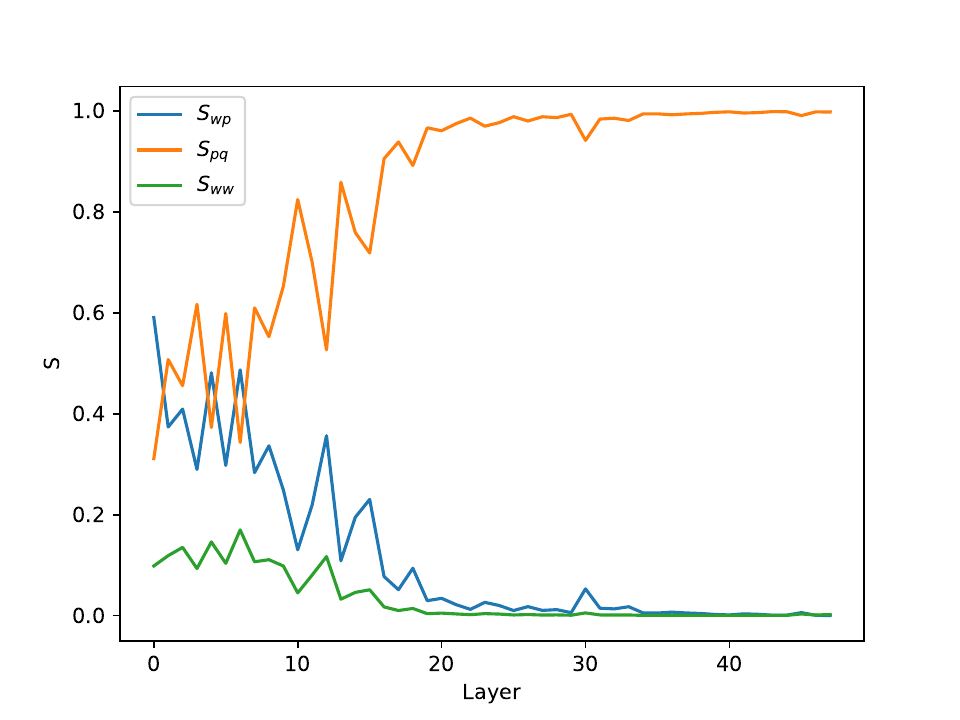}}
\hfill
\subfloat[Results on the AGNews dataset]{
\label{sfig:attn_attr_ag_more_demo}
\includegraphics[height=5cm]{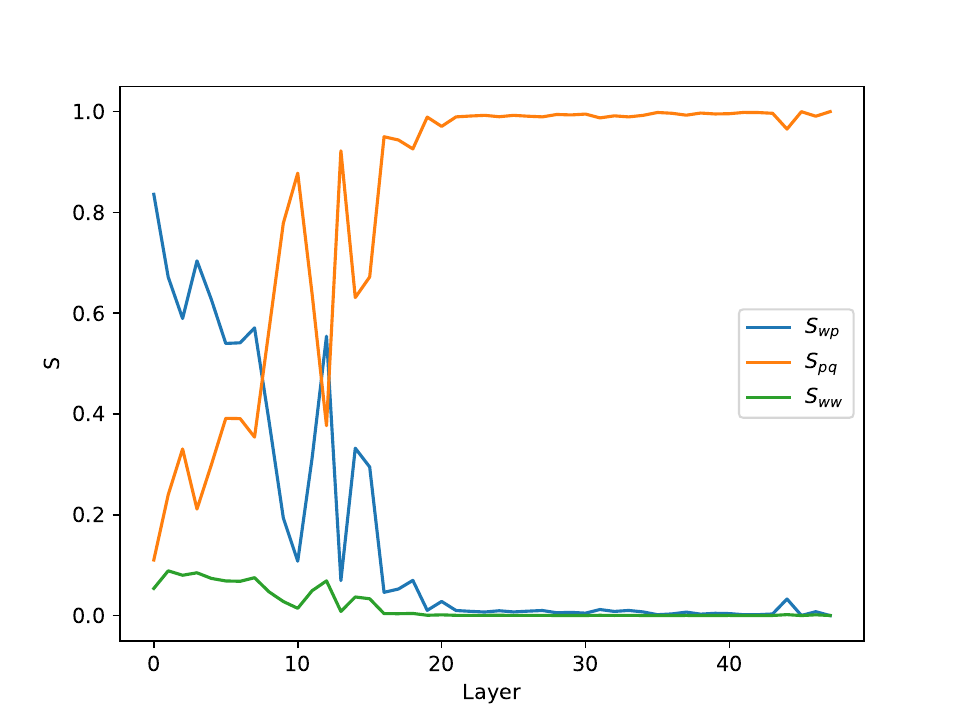}}
% \hspace{2em}
\hfill
\subfloat[Results on the EmoC dataset]{
\label{sfig:attn_attr_emoc_more_demo}
\includegraphics[height=5cm]{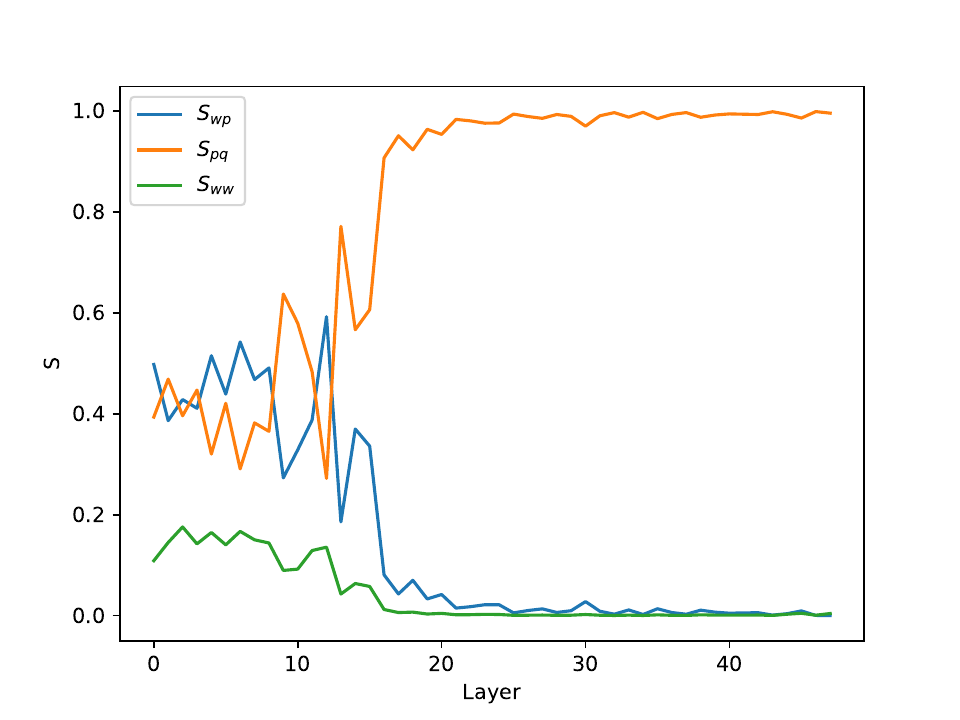}}

\caption{Relative sizes of $S_{wp}$, $S_{pq}$, and $S_{ww}$ when more demonstrations are employed.}
\label{fig:attn_attr_demonstration_more_demo}
\end{figure}

\begin{figure}[t!]
\centering
\subfloat[Effect of different numbers of isolated layers on GPT2-XL]{
\label{fig:aggregation_loyalty_gpt2-xl_more_demo}
\includegraphics[height=5cm]{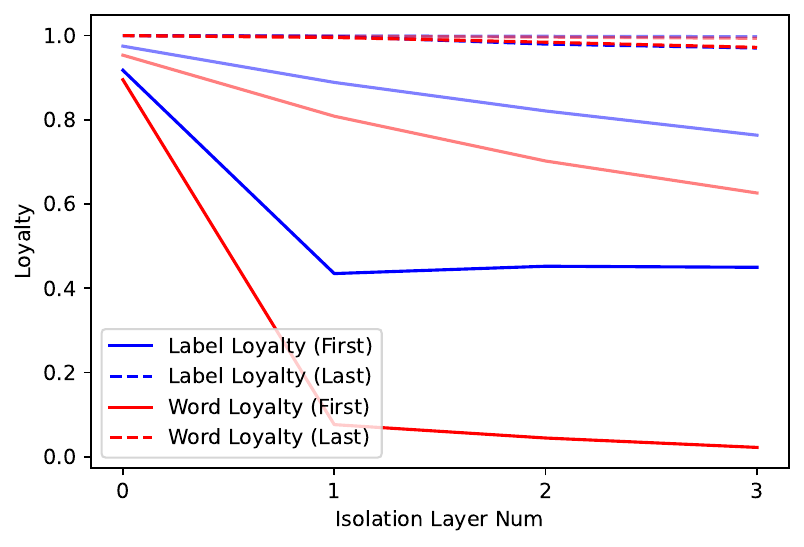}}
% \hspace{2em}
\hfill
\subfloat[Effect of different numbers of isolated layers on GPT-J]{
\label{fig:aggregation_loyalty_gpt-j_more_demo}
\includegraphics[height=5cm]{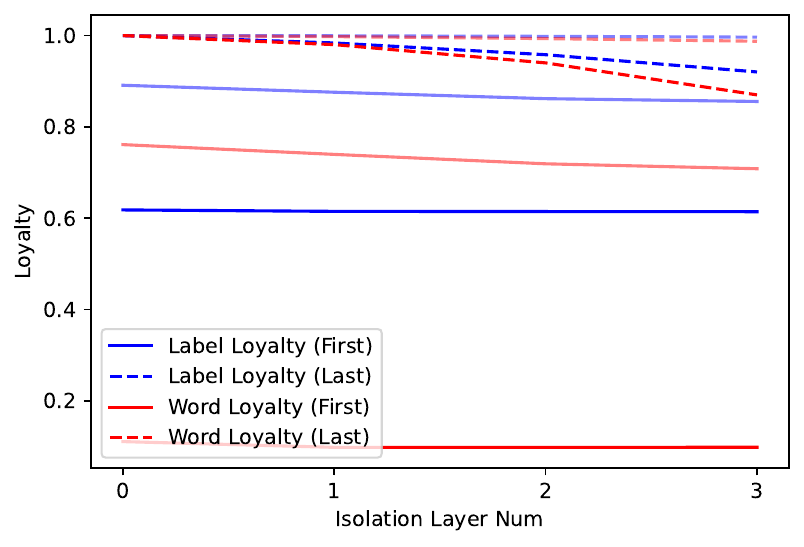}}
\caption{Variations in label loyalty and word loyalty when more demonstrations are employed.}
\label{fig:aggregation_loyalty_more_demo}
\end{figure}

\begin{figure}[t!]
  \centering
  \subfloat[GPT2-XL (total 48 layers).]{
    \includegraphics[width=0.45\textwidth]{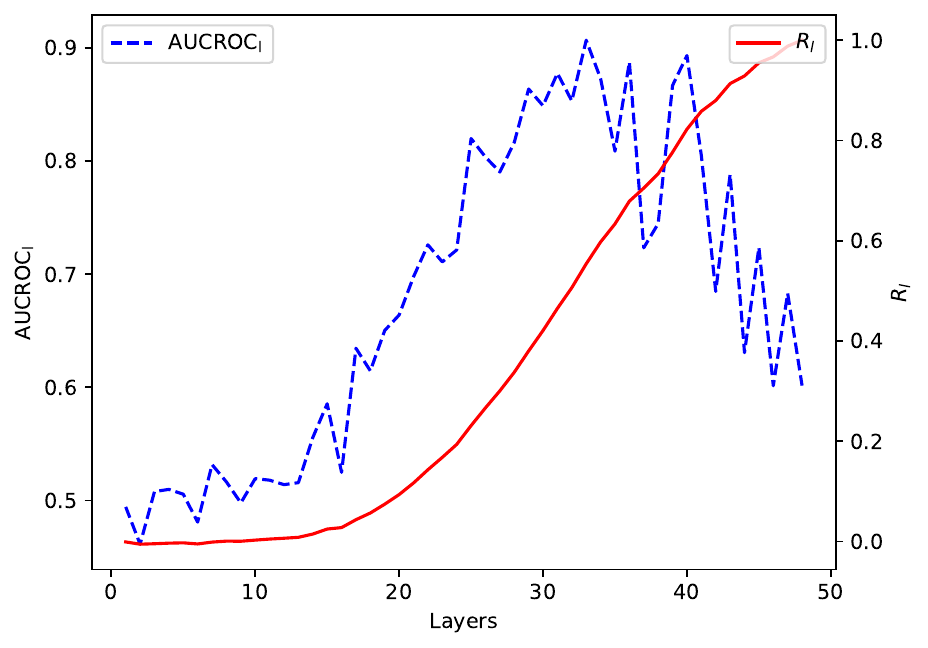}
    \label{fig:auc-roc-ratio-gpt2-xl_more_demo}}
  \hfill
  \subfloat[GPT-J (total 28 layers).]{
    \includegraphics[width=0.45\textwidth]{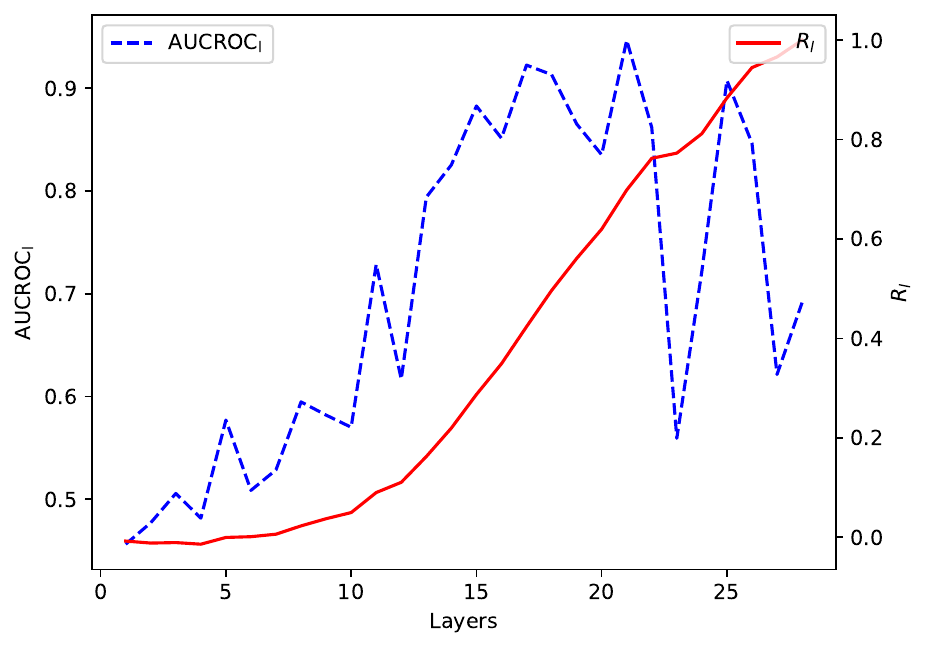}
    \label{fig:auc-roc-ratio-gpt-j_more_demo}}
  \caption{$\text{AUCROC}_l$ and $R_l$ of each layer in GPT models when more demonstrations are employed.
  } 
  \label{fig:auc-roc-ratio_more_demo}
\end{figure}

We implement our experimental analysis utilizing two demonstrations per class, resulting in a total of 4, 12, 8, and 8 demonstrations respectively for SST-2, TREC, AGNews, and EmoC. Our findings, as depicted in Figure~\ref{fig:attn_attr_demonstration_more_demo}, Figure~\ref{fig:aggregation_loyalty_more_demo}, and Figure~\ref{fig:auc-roc-ratio_more_demo}, exhibit a high degree of similarity to the results obtained from experiments that employ one demonstration per class.

\subsection{Results for In-Context Learning with semantically-unrelated labels}
\label{app:SULICL}
\begin{figure}[t!]
\centering
\includegraphics[width=0.45\textwidth]{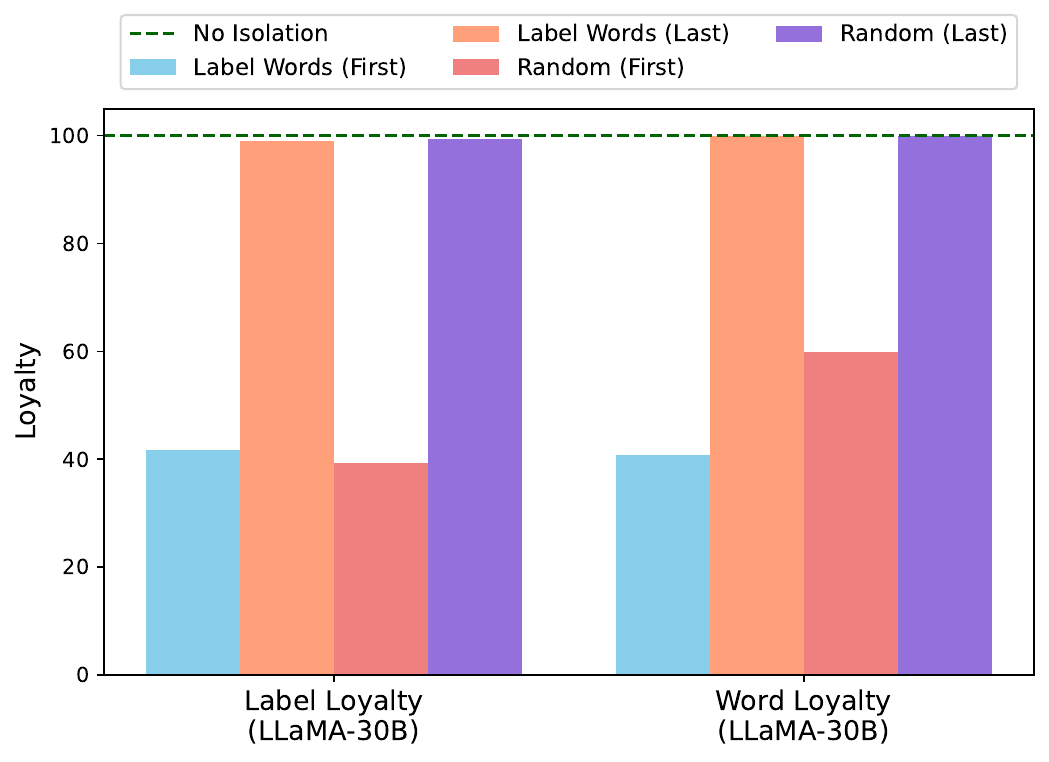}
\caption{The impact of isolating label words versus randomly isolating non-label words within the first or last 5 layers. Isolating label words within the first 5 layers exerts a more pronounced effect, highlighting the importance of shallow-layer information aggregation via label words.
}

\label{fig:llama-1}
\end{figure}

\begin{figure}[t!]
\centering
\includegraphics[width=0.45\textwidth]{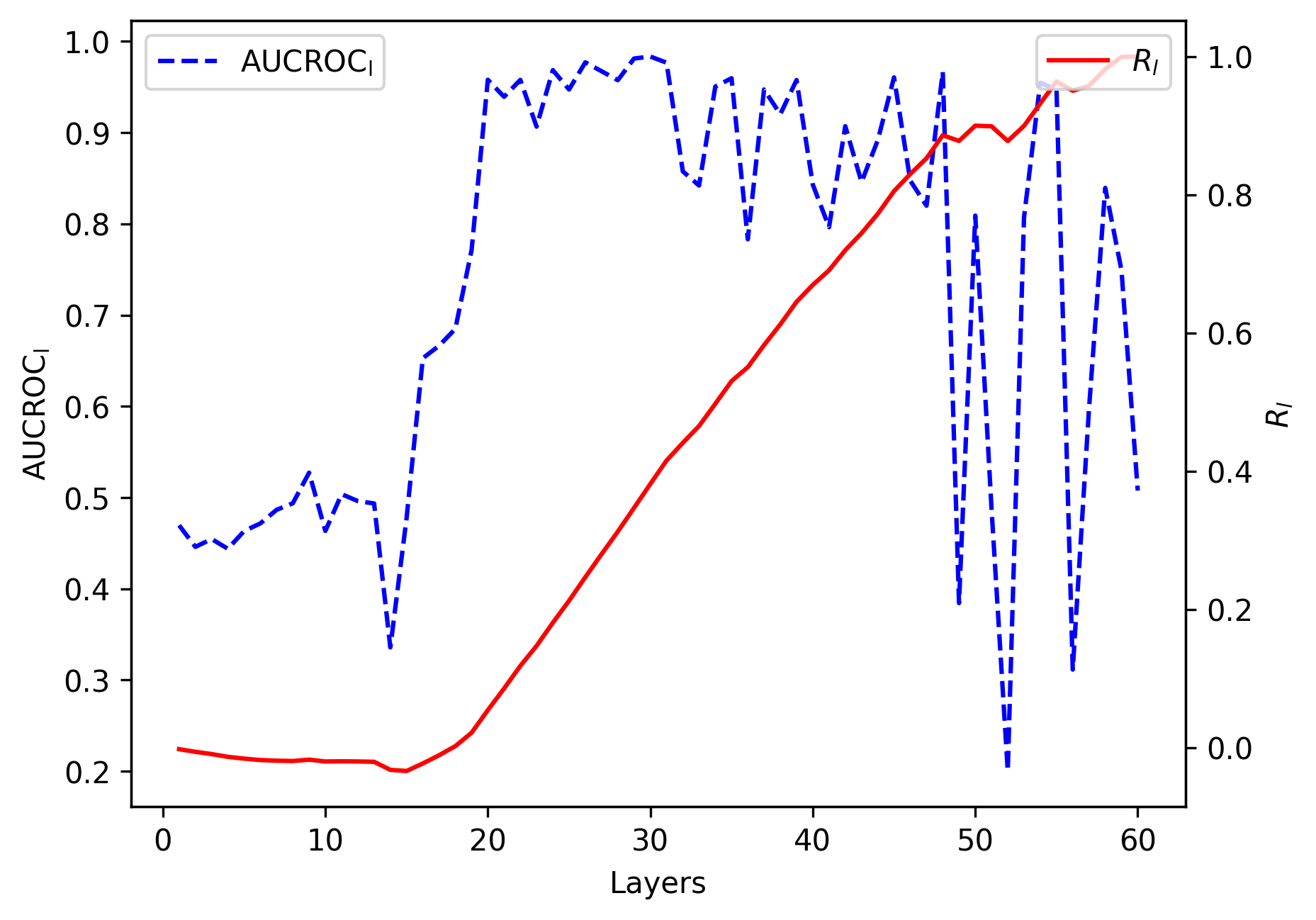}
\caption{AUCROC$_l$ and $R_l$ of each layer of LLaMA-33B on SST-2. Still, deep layers display higher relevance to model prediction, reinforcing the idea that the model extracts information from deep-layer anchors for classification.  
}

\label{fig:llama}
\end{figure}

The applicability of our analytical conclusions to ICL variants, such as the semantically unrelated label ICL~\cite{Wei2023LargerLM}, is an intriguing subject. Given that both GPT2-XL and GPT-J-6B perform at levels akin to random guessing in this ICL setting, we chose LLaMA-33B~\cite{Touvron2023LLaMAOA} and SST-2 for our experiment. We substituted labels with 'A'/'B', and adhered to a similar experimental setup as in sections \S~\ref{subsec:shallow_layer} and \S~\ref{subsec:deep_layer}. However, we applied eight shots per class to facilitate the model in achieving an accuracy of 83.0\% on SST-2. The outcomes align with those derived in \S~\ref{subsec:shallow_layer} and \S~\ref{subsec:deep_layer}. Figure~\ref{fig:llama-1} shows the more pronounced impact of isolating labels in the shallow layers compared to their isolation in the deep layers or the isolation of non-label tokens. Figure~\ref{fig:llama} confirmed that the model leverages information from anchors in the deeper layers to perform classification.

% It is an interesting topic whether our analysis conclusion works for variants of ICL, like semantically-unrelated label ICL~\cite{Wei2023LargerLM}. Since that GPT2-XL and GPT-J-6B perform almost as random guessing in such a ICL setting, we choose LLaMA-33B~\cite{Touvron2023LLaMAOA} and SST-2 to perform the experiment. We replace labels with 'A'/'B', and follow a similar setting like \S~\ref{subsec:shallow_layer} and \S~\ref{subsec:deep_layer}, but with 8 shot per class to ensure the model achieves an accuracy of 83.0\% on this task. The results is consistent with \S~\ref{subsec:shallow_layer} and \S~\ref{subsec:deep_layer}. Specifically, Figure~\ref{fig:llama} shows that isolating labels in the first several layers has a greater impact than isolating in the last several layers or isolating non-labels, and Figure~\ref{fig:llama} affirms the model extracts information from anchors in deep layers to perform classification.

\section{Implementation of Anchor Re-weighting}
\label{appendix:anchor_re-weighting}
In order to implement anchor re-weighting, specific adjustments are made in the model's computational process. After calculating the attention matrix $A_l^h$ of the $h$th head in the $l$th layer, we multiply each $A_l^h(q,p_i)$ by $\exp(\beta^i_{0,lh})$ before proceeding with further computations. This means that for each attention head, we introduce the following modifications:
\begin{equation}
\small 
\begin{aligned}
&\operatorname{Attention}_l^h(Q, K, V)=\hat{A}_l^h V, \\
&A_l^h = \operatorname{softmax}\left(\frac{Q K^T}{\sqrt{d}}\right),\\
&\hat{A}_l^h(k,j) = \begin{cases}
      \exp(\beta^i_{0,lh}) A_l^h(k,j), & \text{if}\ k =q, j=p_i \\
      A_l^h(k,j), & \text{otherwise}
\end{cases}.
\end{aligned}
\end{equation}

\section{Training Settings of Anchor Re-weighting}
\label{appendix: anchor-re-weighting-settings}

For each random seed, we fix the demonstration and sample $1000$ test samples from the test datasets as described in \S~\ref{subsec:shallow_layer}. The optimization of parameter vector $\boldsymbol{\beta}$ is carried out using gradient descent, specifically with the Adam optimizer \cite{Kingma2014AdamAM}. The learning rate is set at $0.01$, with $\beta_1 = 0.9$ and $\beta_2 = 0.999$. Due to memory constraints, we use a batch size of 1. This optimization process is repeated for $10$ epochs. Owing to limitations in computational resources, we restrict our evaluation to the GPT2-XL model and exclude the GPT-J model from our assessment.

\section{The Factor of $L_{\text{demo}}$ and $L_{\mathbf x}$}
\label{appendix:L_demo}
% We discuss the factor of the total length of the demonstrations~$L_{\text{demo}}$ and the length of the text to be predicted~$L_{\mathbf x}$ in compression.
\begin{table}[htb]
\small
  \centering
  \begin{tabular}{c|cccc}
    \toprule
     & SST-2 & TREC & AGNews & EmoC \\
    \midrule
    GPT2-XL & $1.1\times$ & $1.5\times$ & $2.5\times$ & $1.4\times$ \\
    GPT-J & $1.5\times$ & $2.2\times$ & $2.9\times$ & $1.9\times$\\
    % $\frac{L_{\text{demo}}+L_{\mathbf x}}{L_{\mathbf x}}$ & $1.9$ & $9.7$ & $5.1$ & $5.4$ \\
    \midrule
    $L_{\text{demo}}$ & $18$ & $61$ & $151$ & $53$ \\
    $L_{\mathbf{x}}$ & $19$ & $7$ & $37$ & $12$ \\
    \bottomrule
  \end{tabular}
    \caption{Acceleration ratios, $L_{\text{demo}}$ and $L_{\mathbf x}$.}
  \label{tab:compress_eff_v2}
\end{table}
% From Table~\ref{tab:compress_eff_v2}, we can observe that a larger ratio of total length to predicted text length is likely to bring a larger acceleration ratio. And the acceleration ratio of longer demonstrations is higher~(AGNews).
From Table~\ref{tab:compress_eff_v2}, we observe a correlation between the acceleration ratios and the ratio of the total demonstration length ($L_{\text{demo}}$) to the length of the text predicted ($L_{\mathbf{x}}$). It suggests that a greater ratio of total length to predicted text length may yield a higher acceleration ratio.

In addition, the table illustrates that datasets with longer demonstration lengths tend to exhibit higher acceleration ratios. For instance, the AGNews dataset, which has the longest $L_{\text{demo}}$, presents the highest acceleration ratio among the datasets analyzed. These findings could indicate an increased efficiency of the Hidden$_\text{anchor}$ method in contexts involving longer demonstration lengths.

\section{Calculation of $\hat{\mathbf{k}}$}
\label{appendix:hat_k}
For the sampled sequence $x_1,...,x_T$ to be predicted, we denote the query vectors of the target positions as $\mathbf q_1,...,\mathbf q_T$. We then compute the matrix  $\hat{\mathbf{Q}} = (\mathbf q_1 - \overline{\mathbf{q}},...,\mathbf q_T-\overline{\mathbf{q}})$ by subtracting the mean vector, $\overline{\mathbf{q}}$, from each query vector.
Subsequently, we determine the $M$ directions, $\mathbf{v}_1,...,\mathbf{v}_M$, that correspond to the M largest variation directions for the centralized query vectors $\hat{\mathbf{q}}_1,...,\hat{\mathbf{q}}_T$. The $i^{th}$ direction, $\mathbf{v}_i$, is chosen to maximize the variance of the projection of the centralized query vectors onto it, while also being orthogonal to the previously chosen directions, $\mathbf{v}_1,...,\mathbf{v}_{i-1}$. This process can be formalized as follows:
\begin{equation}
    \begin{aligned}
    & \mathbf{v}_1=\underset{\|\mathbf{v}\|=1}{\arg \max } \operatorname{Var}\left\{\mathbf{v}^{\top} \hat{\mathbf{Q}}\right\},  \\
    & \mathbf{v}_2=\underset{\|\mathbf{v}\|=1, \mathbf v \perp \mathbf v_1}{\arg \max } \operatorname{Var}\left\{\mathbf{v}^{\top} 
    \hat{\mathbf{Q}}\right\},  \\
    & ...\\
    &  \mathbf{v}_M=\underset{\|\mathbf{v}\|=1, \mathbf v \perp \mathbf v_1,...,\mathbf v \perp \mathbf v_{M-1}}{\arg \max } \operatorname{Var}\left\{\mathbf{v}^{\top} 
    \hat{\mathbf{Q}}\right\}.
\end{aligned}
\end{equation}

We define $\sigma_i$ as the square root of the variance of the projection of $\hat{\mathbf{Q}}$ onto the $i^{th}$ direction, i.e., $\sqrt{\operatorname{Var}\left\{\mathbf{v}_i^{\top} \hat{\mathbf{Q}}\right\}}$.

To derive features $\hat{\mathbf{k}}$s, we project the key vector $\mathbf{k}$ onto the directions $\mathbf{v}_1,...,\mathbf{v}_M$ and scale the projections by the corresponding standard deviations $\sigma_1,...,\sigma_M$. Each feature, $\hat{\mathbf{k}}_i$, is thus calculated as $\sigma_i\mathbf{v}_i^T\mathbf{k}$.

\begin{figure*}[t!]
\centering
\subfloat[$M=5$]{
    \label{sfig:M5}
    \includegraphics[height=5cm]{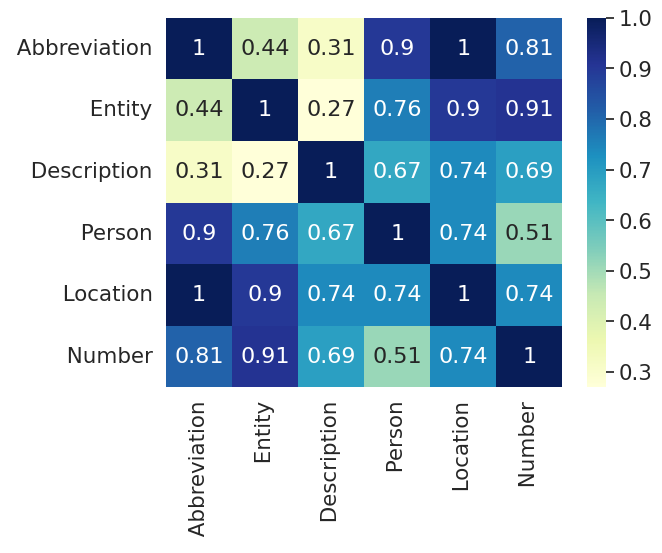}}
    % \vspace{\baselineskip}
    % \hfill
    \hspace{2em}
  \subfloat[$M=10$]{
    \label{sfig:M10}
    \includegraphics[height=5cm]{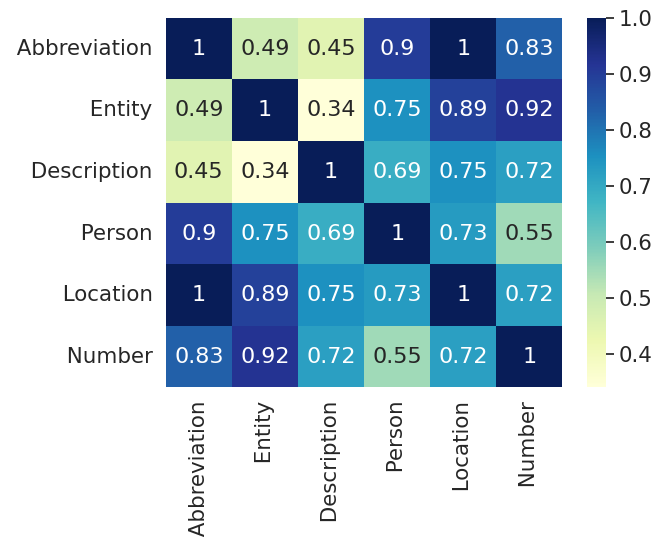}}
% \vspace{\baselineskip}
\par
    
\subfloat[$M=20$]{
    \label{sfig:M20}
    \includegraphics[height=5cm]{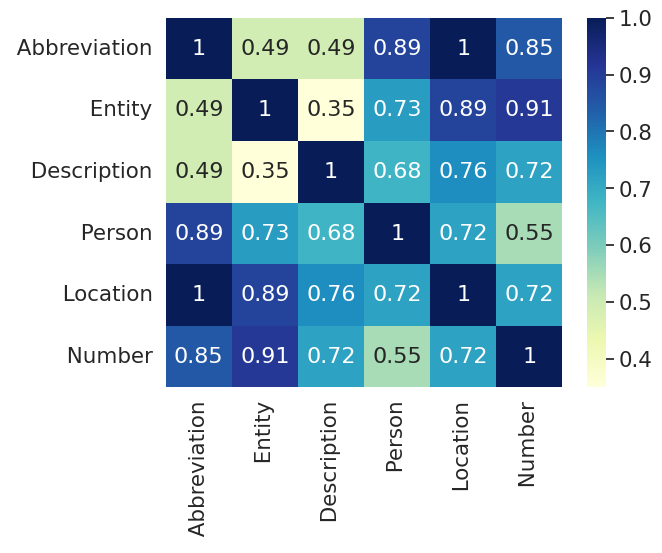}}
% \caption{Predicted confusion matrices under $M = 5,10,20,50,100,200$.}
% \end{figure}
% \hfill
% \begin{figure}[t!]
%     \ContinuedFloat % 继续上一页的 figure
%     \centering
\hspace{2em}
\subfloat[$M=50$]{
    \label{sfig:M50}
    \includegraphics[height=5cm]{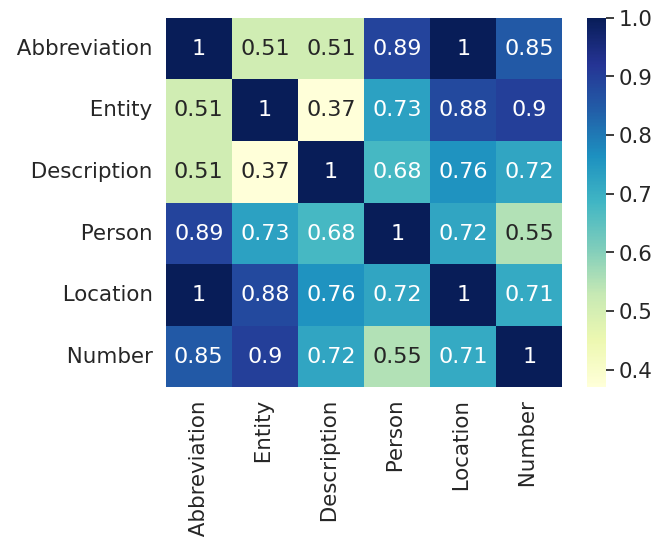}}

   \vspace{\baselineskip}
    % \vspace{\baselineskip}
\subfloat[$M=100$]{
    \label{sfig:M100}
    \includegraphics[height=5cm]{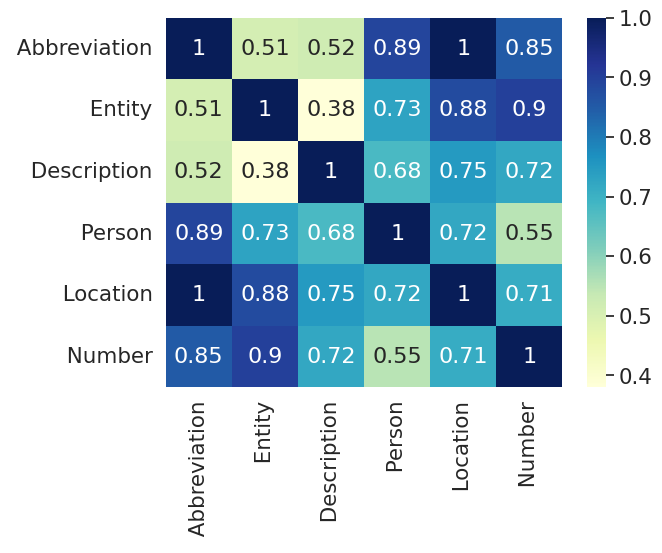}}
% \vspace{\baselineskip}
% \hfill
\hspace{2em}
\subfloat[$M=200$]{
    \label{sfig:M200}
    \includegraphics[height=5cm]{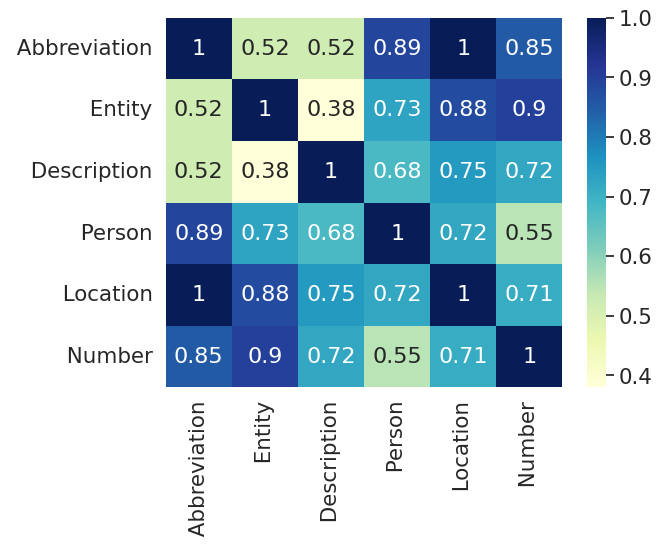}}
    
  \caption{Predicted confusion matrices under $M = 5,10,20,50,100,200$.}
  \label{fig:conf_M}

% attention_attr_ana.ipynb
\end{figure*}

We further examine the influence of $M$ on the prediction confusion matrix, $\text{Confusion}{ij}^{\text{pred}}$, as depicted in Figure~\ref{fig:conf_M}. Given the similarity in outcomes for various $M$, we settle on a value of $M=10$ for computation of $\text{Confusion}{ij}^{\text{pred}}$.

\section{Calculation of $\text{Confusion}_{ij}$}
\label{appendix:confusion}
To gauge the true degree of confusion between categories $i$ and $k$ for a given model, we suggest utilizing the $\text{Confusion}_{ij}$ metric:

First, we procure all test samples $x_t$ bearing true labels $i$ or $k$. We then obtain the probabilities $p_i^t$ and $p_j^t$ yielded by the model for categories $i$ and $k$, respectively, on these samples. These probabilities are normalized to a total of 1. Essentially, we derive a classifier $f$ that delivers the probabilities $p_i^t$ and $p_j^t$ for the categories $i$ and $k$ respectively, on the test samples $x_t$. By calculating the Area Under the Receiver Operating Characteristic Curve (AUC-ROC) value of this classifier $f$, we get the degree of confusion between category $i$ and $k$, termed as $\text{Confusion}_{ij}$.

The computed $\text{Confusion}{ij}$ is a value that never exceeds 1. The closer $\text{Confusion}{ij}$ approximates 1, the less pronounced the confusion, and vice versa.

We use the above metric instead of directly analyzing the output labels of the model because previous work has indicated the issue of insufficient output probability calibration in ICL~\cite{Zhao2021CalibrateBU}, which is greatly affected by factors such as sample ordering and model preferences for specific label words. By leveraging our defined degree of confusion, $\text{Confusion}_{ij}$, we can implicitly alleviate the disturbances arising from insufficient probability calibration on the output labels. This allows for a more accurate representation of the model's degree of confusion for different categories, mitigating the impact of randomness.
% \section{Effect of different $M$ on the predicted confusion matrix}
% \label{app:conf_M}

% We plot the predicted confusion matrix under $M = 5,10,20,50,100,200$, and find that the choice of M does not significantly influence the predicted confusion matrix. (shown in figure~\ref{fig:conf_M})

\section{Reproducibility}
In the supplementary material, we have provided codes that allow for the faithful replication of our experiments and subsequent result analysis. To ensure consistency and reproducibility across different devices, we have fixed the five random seeds to the values of 42, 43, 44, 45, and 46. We invite readers to delve into the code for additional implementation details that may arouse their interest.

\end{document}